\documentclass[journal]{IEEEtran}
\pdfoutput=1

\usepackage{hyperref}
\usepackage{graphicx}
\usepackage{comment}
\usepackage{amssymb}
\usepackage{subcaption}
\usepackage{array}
\usepackage{url}
\usepackage[export]{adjustbox}
\usepackage{multirow}
\usepackage{tabularx,booktabs}
\usepackage{caption}
\usepackage[justification=centering]{caption}
\usepackage{amsmath}

\begin{document}



\title{Multiclass Wound Image Classification using an Ensemble Deep CNN-based Classifier}




\author{Behrouz~Rostami\textsuperscript{*}, D.M.~Anisuzzaman, Chuanbo~Wang, Sandeep~Gopalakrishnan, Jeffrey~Niezgoda, and~Zeyun~Yu\textsuperscript{*}
\thanks{Behrouz Rostami is with the Electrical Engineering Department, University of Wisconsin-Milwaukee, Milwaukee, WI, 53211 USA.}
\thanks{D. M. Anisuzzaman, Chuanbo Wang, and Zeyun Yu are with the Computer Science Department, University of Wisconsin-Milwaukee, Milwaukee, WI, 53211 USA.}
\thanks{Sandeep Gopalakrishnan is with the College of Nursing, University of Wisconsin-Milwaukee, Milwaukee, WI, 53211 USA.}
\thanks{Jeffrey Niezgoda is with the AZH Wound Center, Milwaukee, WI, 53221 USA.}
\thanks{Corresponding authors: brostami@uwm.edu (Rostami); yuz@uwm.edu (Yu)}}


\markboth{}
{Shell \MakeLowercase{\textit{et al.}}: Bare Demo of IEEEtran.cls for IEEE Journals}

\maketitle

\begin{abstract}
Acute and chronic wounds are a challenge to healthcare systems around the world and affect many people's lives annually. Wound classification is a key step in wound diagnosis that would help clinicians to identify an optimal treatment procedure. Hence, having a high-performance classifier assists the specialists in the field to classify the wounds with less financial and time costs. Different machine learning and deep learning-based wound classification methods have been proposed in the literature. In this study, we have developed an ensemble Deep Convolutional Neural Network-based classifier to classify wound images including surgical, diabetic, and venous ulcers, into multi-classes. The output classification scores of two classifiers (patch-wise and image-wise) are fed into a Multi-Layer Perceptron to provide a superior classification performance. A 5-fold cross-validation approach is used to evaluate the proposed method. We obtained maximum and average classification accuracy values of 96.4\% and 94.28\% for binary and 91.9\% and 87.7\% for 3-class classification problems. The results show that our proposed method can be used effectively as a decision support system in classification of wound images or other related clinical applications.
\end{abstract}




\section{Introduction}
\label{S:1}
\IEEEPARstart{A}{cute} and chronic wounds are a challenge and burden to healthcare systems worldwide. In the United States alone, acute wounds affect 11 million people and chronic wounds influence more than 6 million humans annually with an estimated medicare burden of \$28.1 billion to \$96.8 billion (US)~\cite{demidova2012acute,sen2009human,sen2019human}. The characterization of a wound is a key step in wound diagnosis that would help clinicians to identify an optimal treatment procedure and assess the efficacy of the treatment. Wound classification is to classify a wound as a whole into different types (e.g., venous, diabetic, pressure) or different conditions (e.g., ischemia vs. non-ischemia, infection vs. non-infection), which constitutes an essential part of wound characterization and assessment.\par
Wounds are traditionally manually classified by wound specialists and made as part of the electronic health record (EHR). With new developments of Artificial Intelligence (AI) in the past decades, however, intelligent algorithms have been popularly used in healthcare in such fields as drug discovery, eye care, medical image diagnostic systems, etc.~\cite{yu2018artificial} in addition to the wound classification under investigation in the current study. In recent years, AI algorithms have evolved into so-called data-driven techniques without human or expert intervention as opposed to the early generations of AI that were rule-based relying largely on an expert's knowledge~\cite{yu2018artificial}. Radiology, ophthalmology, immunology, genetics, and wound care are just a few examples of using developed AI algorithms like Machine Learning (ML) methods in health care~\cite{yu2018artificial,lakhani2018machine,figgett2019machine,andreatta2017machine,bari2017machine,rahman2018machine,collier2019lotus}. \par
A representative of the data-driven developments is Deep Learning (DL), which is able to analyze complicated data automatically and extract the needed information, relationships, and patterns~\cite{ohura2019convolutional,lakhani2018machine}. DL models may appear in different forms such as Deep Convolutional Neural Networks (DCNN), according to the utilized building blocks in their structures. DCNNs based on deep learning theory, are neural networks with multiple hidden layers unlike their ancestors which utilized a limited number of layers~\cite{jiang2017artificial}. In these networks, convolution is used as the main math operation to process the input information~\cite{rostami2019survey}.
   Many studies have been conducted in the wound care field using DL~\cite{wang2015unified,li2018composite,rajathi2019varicose} and DCNNs, specifically in wound image analysis tasks like segmentation~\cite{goyal2019skin,veredas2015wound} and classification~\cite{abubakar2019discrimination,zahia2018tissue,zhao2019fine}. These studies have clearly shown the effectiveness and efficiency of deep convolutional neural networks in wound diagnosis and analysis.

\begin{figure}[h!]

\centering\includegraphics[width=8.8cm,height=3.5cm]{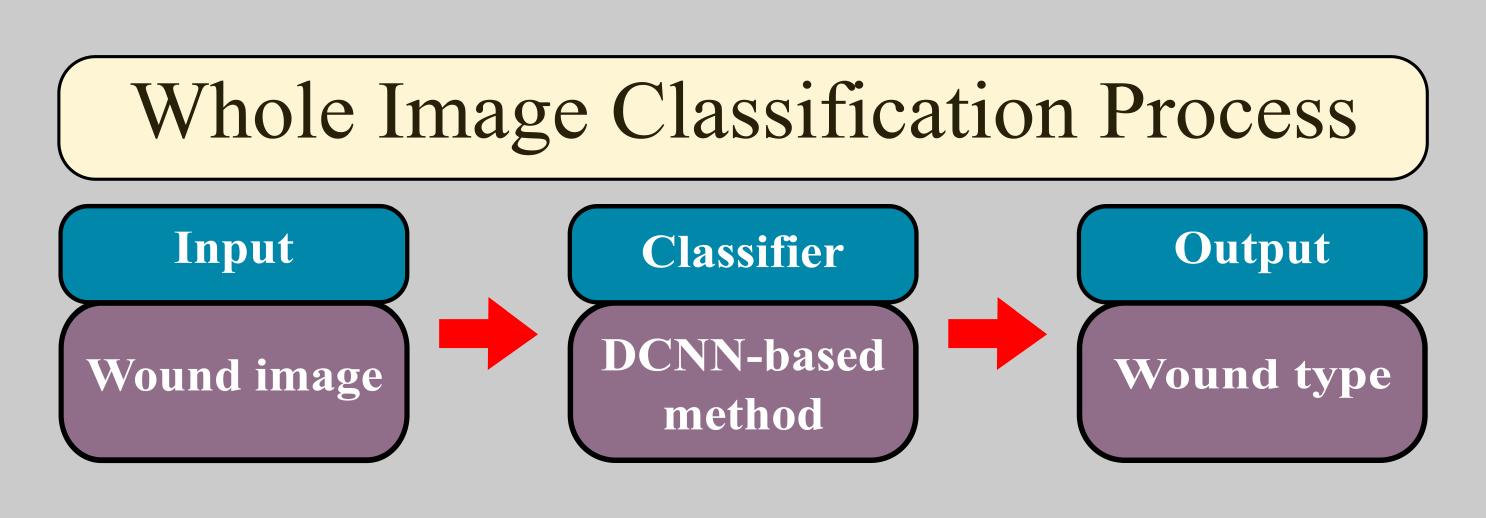}

\caption{Classification process.}
\label{fig:goal}
\end{figure}

The main contribution of this research is to propose an end-to-end ensemble DCNN-based classifier for classifying the entire wound images into multiple wound types. To the best of our knowledge, this research is the first study in which an ensemble method used for image-wise classification of the wound images into different types. 
We have proposed an innovative combining strategy to combine a sliding window-based patch classifier and a common DCNN, AlexNet to classify the wound images. Moreover, far as we are aware, this is the first time a deep learning-based method is proposed for classifying the wound images into more than two types. By accepting the entire wound image as the input, our proposed approach is able to generate the wound type as the output. Figure~\ref{fig:goal} displays an overview of our proposed method. Additionally, we utilized a new dataset of real wound images containing image data with more wound types than those considered in the prior publications. Our dataset contains 538 images from four different types of wounds including diabetic, venous, pressure, and surgical wounds. All these images along with their types and corresponding extracted ROIs will be made publicly available.\par
The rest of this manuscript has been organized as follows: Section~\ref{S:2} gives a literature review. In Section~\ref{S:3}, we present the materials used and discuss the details of our proposed ensemble deep CNN-based classification method. Section~\ref{S:4} describes the experimental results. We give the summary and discussion of the results in Section~\ref{S:5}, followed by our conclusions in Section~\ref{S:6}.

\section{Related works}
\label{S:2}
\begin{figure*}[h!]

\centering\includegraphics[width=0.9\linewidth]{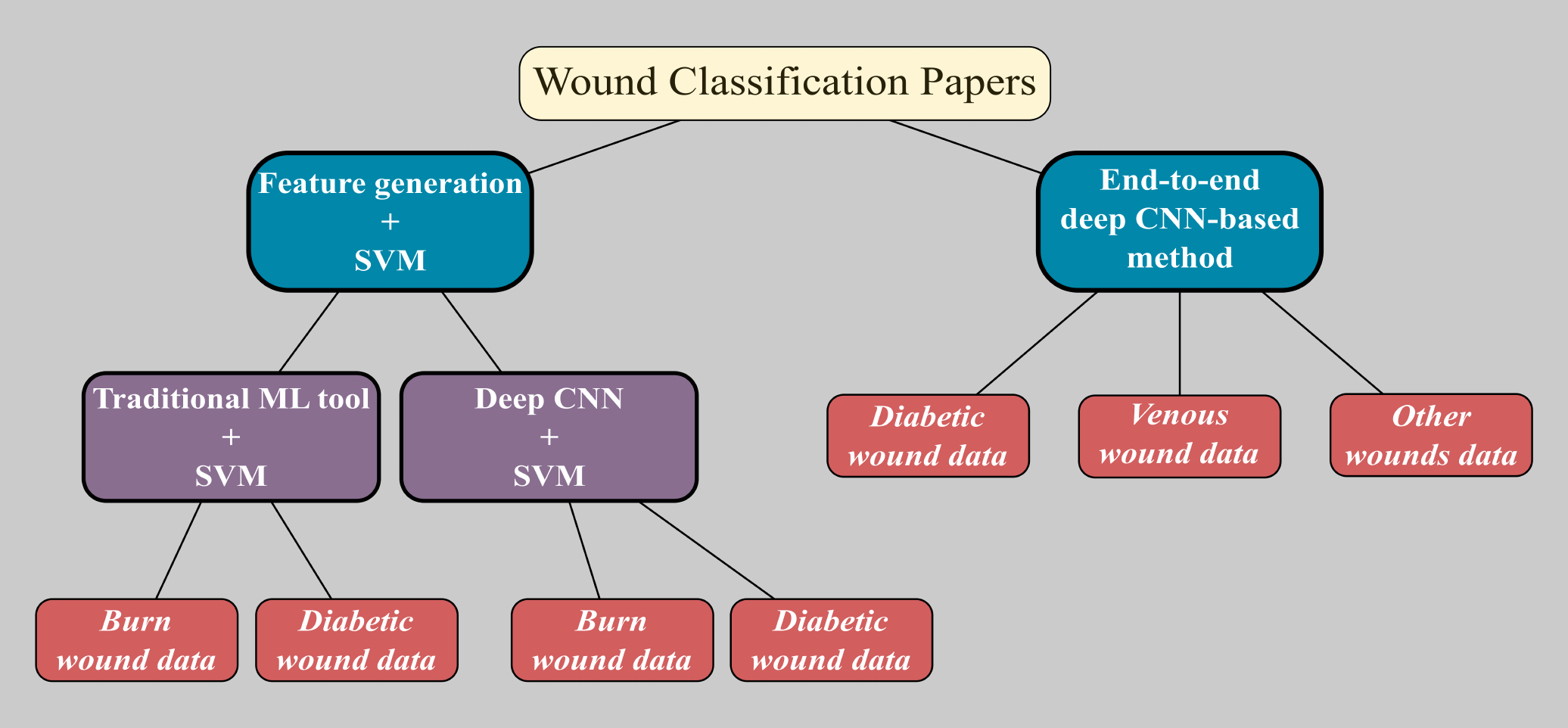}

\caption{Organization chart for the wound classification papers.}
\label{fig:chart}
\end{figure*}

In this section, we review the previous wound image classification studies. We have organized the articles under two major subcategories: the papers which used a feature extraction method followed by an SVM, and the articles in which an end-to-end DCNN-based classification approach was proposed. The complete organization chart for the reviewed studies can be found in Figure~\ref{fig:chart}. 

\subsection{Feature generation + SVM}
\subsubsection{Traditional ML algorithm + SVM}
Yadav et al. proposed a method for binary classification of burn wound images using machine learning tools~\cite{yadav2019feature}. The authors classified the images into the two categories, grafting and non-grafted wounds, using a classic color-based feature extraction approach followed by an SVM. The dataset included 94 images with different burn depths including full-thickness, deep dermal, and superficial dermal. A classification accuracy of 82.43\% was reported. Testing the proposed method on a very small dataset is the weakness of this research.

Goyal et al. suggested detection and localization methods for Diabetic Foot Ulcer (DFU) on mobile devices~\cite{goyal2018robust}. For the classification part, they tried both conventional and DCNN-based methods. A dataset with 1775 DFU images was used and the ground truth generated by creating bounding boxes around the ROI using an annotation software. For the traditional machine learning techniques, patches were extracted from normal skin and abnormal areas and the number of samples increased by utilizing data augmentation methods. Different traditional feature extraction algorithms were applied, and the best three methods were selected. Then they extracted 209 features from each patch which were used for training a Quadratic SVM classifier. Finally, for a new image, sliding window technique was used for classifying each patch of the image as normal or abnormal by utilizing the trained SVM.

\subsubsection{Deep CNN + SVM}
In another study, Abubakar et al. proposed a machine learning-based approach to distinguish between burn wounds and pressure ulcers~\cite{abubakar2019can}. Pre-trained deep architectures like VGG-face, ResNet101, and ResNet152 were utilized for feature extraction. The features were fed into an SVM for the classification task. The dataset included 29 pressure and 31 burn images which were augmented using cropping, rotation, and flipping transformations. After augmentation, they performed binary and 3-class classification experiments. In binary classification experiment, the images were classified into burn or pressure categories and in 3-class classification problem the goal was to classify the images into the labels burn, pressure, or healthy skin. ResNet152 architecture generated the best results for both classification problems with an accuracy value of 99.9\%.\par

Goyal et al. predicted the presence of infection or ischemia in DFUs using a deep learning-based classification method~\cite{goyal2020recognition}. A new dataset with 1459 DFU images was introduced and the samples were augmented using Faster-RCNN and InceptionResNetV2 networks. Binary classification experiments were performed to classify the samples into infection or non-infection, and ischemia or non-ischemia classes. In more detail, some color-based descriptors were extracted from each patch before classification. ResNet50, InceptionV3, and InceptionResNetV2 architectures were used in this study. Besides, the authors used an ensemble CNN approach for combining the outputs of the three deep networks and fed it into an SVM for classification. They used MATLAB and TensorFlow frameworks. In both binary classification problems, the deep learning-based methods showed a better performance than the traditional classifiers. The authors reported the accuracy values of 90\% for ischaemia and 73\% for infection experiments.

\subsection{End-to-end deep CNN-based methods}
In~\cite{goyal2018robust}, for the deep learning-based classification methods, two-tier transfer learning approach was utilized for training the deep architectures including MobileNet, InceptionV2, ResNet101, and InceptionResNetV2. This method uses both partial and full transfer learning which means transferring only the lower level features or the whole features from a pre-trained model to the new model. Tensorflow was used as the framework. Object localization algorithms like R-FCN and Faster R-CNN were utilized followed by the trained deep architectures for tasks like classification. The combination of Faster R-CNN and InceptionV2 reported as the best model.\par
In another research, Goyal et al. used convolutional neural networks to classify diabetic foot ulcers~\cite{goyal2018dfunet}. A DFU image dataset with 397 images was presented. Data augmentation techniques were utilized to increase the number of samples. They proposed DFUNet, a deep neural network, for patch-wise classification of the foot ulcers into either normal or abnormal classes. DFUNet utilized the idea of concatenating the outputs of three parallel convolutional layers which used different filter sizes. The authors claimed that using this idea, multiple-level features were extracted from the input which resulted in having a network with higher discriminative strength. An accuracy value of 92.5\% was reported for the proposed method. The main issue about this research is that the authors proposed a patch classifier which is not very helpful in medical image classification tasks. Indeed, it makes more sense to the clinicians and is more useful, to work with a whole-image classifier instead of a patch classification model. \par
 Nilsson et al. proposed a CNN-based method for venous ulcer image classification~\cite{aguirre2018classification}. The utilized dataset included 300 samples and a VGG-19 network was used to classify the images into venous or non-venous categories. The methodology included pre-training of the VGG-19 network using another dataset of Dermoscopic images and then, fine-tuning the network utilizing their related dataset.  The values obtained for accuracy, precision, and recall reported as 85\%, 82\%, and 75\%, respectively. Caffe, TensorFlow, and keras were used as the frameworks.\par
 
 Alaskar et al. applied deep convolutional neural networks for intestinal ulcer detection in wireless capsule endoscopy images~\cite{alaskar2019application}. AlexNet and GoogleNet architectures were utilized to classify the input images into ulcer (abnormal) or non-ulcer (normal) categories. The dataset consisted of 1875 images obtained from wireless capsule endoscopy video frames and the experiments implemented in MATLAB environment. The authors reported classification accuracy of 100\% for both networks. \par
 
 In another research, Shenoy et al. proposed a method to classify wound images into multiple classes using deep CNNs~\cite{shenoy2018deepwound}. A dataset with 1335 wound images collected via smartphones as well as the internet, was used in this study. After pre-processing and augmentation, nine different labels were created and for each label two positive and negative subcategories were considered. The authors created a modified form of VGG16 network, WoundNet, and three different versions of WoundNet were pre-trained on the ImageNet dataset. Besides, another network named Deepwound, which was an ensemble model was designed for averaging of the outcomes from the three individual WoundNet architectures. The algorithms were implemented in Keras. Also, an application was created for mobile phones to facilitate patient to physician consultation and wound healing evaluation. \par

 Alzubaidi et al. presented a DCNN for binary classification of diabetic foot ulcers~\cite{alzubaidi2019dfu_qutnet}. A new dataset consisting of 754 smartphone-captured foot images was introduced in this study. The goal was to classify the samples into normal or abnormal (DFU) skin categories. Normal and abnormal patches were extracted from the images and number of samples increased using data augmentation techniques. The proposed network, DFU\_QUTNet, is a deep architecture with 58 layers including 17 convolutional layers. In comparison with the common DCNNs, the width of the proposed model has been increased without adding computational complexities. Then the network would be able to extract more information from the input which results in higher classification accuracy. In one experiment, DFU\_QUTNet was applied for an end-to-end classification task and in another one, it was utilized as a feature extractor along with SVM and KNN classifiers. The maximum reported F1-Score was 94.5\% obtained from combining DFU\_QUTNet and SVM. Although designing a high-performance patch classifier can be a good achievement, but  in clinical environments it would be more useful to have a whole image classification system and it is the weakness of this research.

\begin{table*}[]
\centering
\footnotesize
\caption{\\Summary of Wound Image Classification Works.}
\label{tab:relatedworks}
\resizebox{\textwidth}{!}{%

\begin{tabular}{|m{1cm}|m{2.5cm}|m{2.5cm}|m{2.3cm}|m{2.5cm}|m{5cm}|}
\hline
\textbf{Ref} & \textbf{Classification} & \textbf{Feature(s)} & \textbf{Methods} & \textbf{Dataset} & \textbf{Limitation(s)} \\ \hline
\cite{goyal2018dfunet}  & Binary (DFU/normal skin). & N/A &          A novel CNN architecture named DFUNet. &    DFU dataset with 397 images (292 abnormal, 105 normal cases). &     For small DFUs and DFUs having similar color like surrounding skins is hard to classify by this network. This also goes for normal skins with wrinkle and high red tones.  \\ \hline

\cite{yadav2019feature}  &      Binary (graft/non-graft burn wound)  & color, texture, and depth.  &  Support Vector Machine (SVM). &    Burns BIP\_US Database. & Very small evaluation set (74 images) was used. \\ \hline

\cite{aguirre2018classification} & Binary (venous/non-venous) &   N/A &   A pre-trained VGG-19 network.  & A dataset with 300 images with specialist annotation.  &  Classification accuracy depends on camera distance from the ulcer surface.\\ \hline

 \cite{alaskar2019application}    &    Binary (ulcer/non-ulcer endoscopy image) &            N/A              &        AlexNet and GoogleNet.               &        1875 images obtained from wireless capsule endoscopy video frames.          &    An unbalanced test set (3:1 ratio) has been used. \\ \hline

        \cite{shenoy2018deepwound}                 &           Binary (considered pos/neg cases for different labels like wound, infection, granulation, etc.
        &       N/A                   &          WoundNet (modified version of VGG-16) and Deepwound (an ensemble model).             &   1335 wound images collected via smartphones and internet.          &        As accuracy varies from 72\% to 97\% for different binary classes, for some specific classes (like: drainage (72\%)) this model does not work well.                  \\ \hline

          \cite{abubakar2019can}    &   Binary (burn/pressure), 3-class (burn/pressure/healthy skin)   &    Feature extracted from VGG-face, ResNet101, and ResNet152.                 &   Support Vector Machine (SVM).     &   29 pressure and 31 burn images.     &     Very small dataset used.  \\ \hline

         \cite{alzubaidi2019dfu_qutnet}        &           Binary (normal/abnormal skin (diabetic ulcer)).              &            N/A              &           A novel deep CNN (DFU\_QUTNet).            &         754 foot images.         &            Only use precision, recall and f1-score as evaluation matrices, which may not reflect all the evaluations clearly.            \\ \hline

         \cite{goyal2020recognition}           &        Binary (infection/non-infection, ischemia/non-ischemia)                 &      Bottleneck features extracted from Inception-V3, ResNet50, and
InceptionResNetV2.    &           Support Vector Machine (SVM).            &         1459 DFU images.         &          Depending on lighting conditions (shadow), marks, and skin tone their model can show poor performance.              \\ \hline

       \cite{goyal2018robust}                  &          Binary (normal/abnormal skin (DFU))               &          209 features extracted using LBP, HOG, and color descriptors.                &           Quadratic SVM, InceptionV2, MobileNet, ResNet101, Inception-ResNetV2            &   1775 DFU images collected from a hospital within five years.        &              No evaluation of their classification task has been given.          \\ \hline
\end{tabular}
}
\end{table*}

Table~\ref{tab:relatedworks} summarizes the reviewed studies. Only a few papers were identified that discuss wound analysis from the wound type classification point of view. Also, most of the publications on wound type classification, discuss only the binary classification problems such as classifying the samples into normal and abnormal cases. Having difficulties to access a reliable dataset can be mentioned as a reason for this issue. Providing data to fill this gap in the literature was one of the motivations for our research. Moreover, many papers discussed only the patch-wise or ROI classification instead of the image-wise wound classification. In the rest of this paper, we propose an ensemble classifier for image-wise multi-class wound type classification using deep convolutional neural networks.

\section{Materials and Methods}
\label{S:3}

\subsection{Dataset}
\begin{figure*}[htp]
\centering
\includegraphics[width=2.7cm,height=3cm]{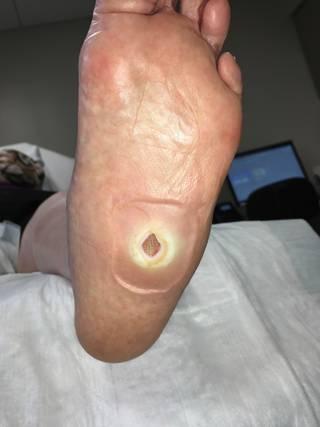}\quad
\includegraphics[width=2.7cm,height=3cm]{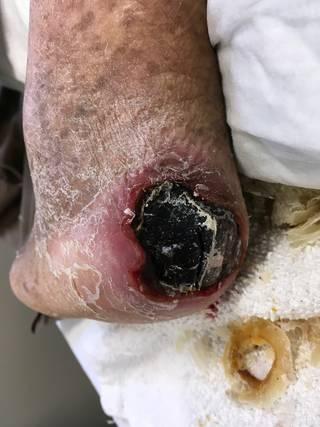}\quad
\includegraphics[width=2.7cm,height=3cm]{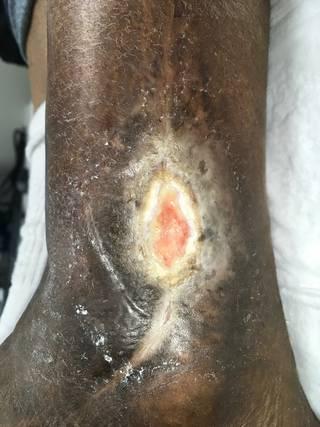}\quad
\includegraphics[width=2.7cm,height=3cm]{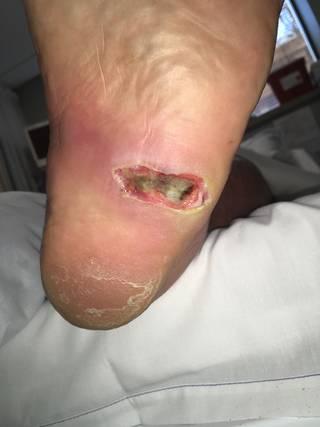}

\medskip

\includegraphics[width=2.7cm,height=3cm]{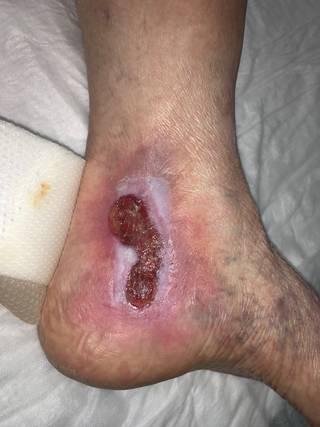}\quad
\includegraphics[width=2.7cm,height=3cm]{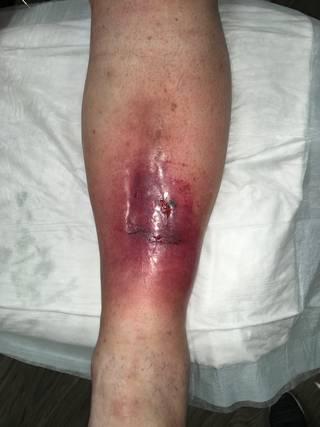}\quad
\includegraphics[width=2.7cm,height=3cm]{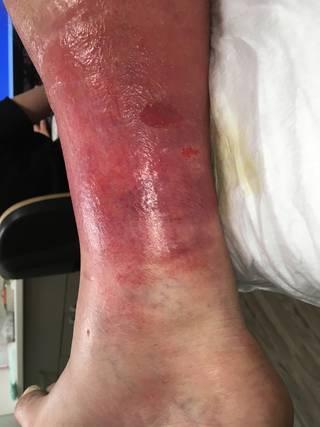}\quad
\includegraphics[width=2.7cm,height=3cm]{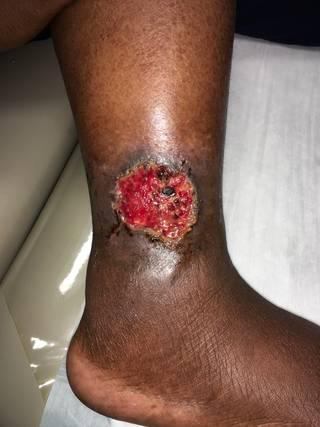}

\medskip

\includegraphics[width=2.7cm,height=3cm]{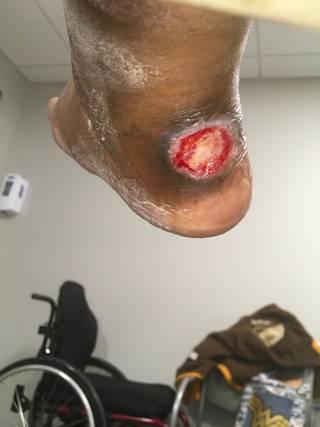}\quad
\includegraphics[width=2.7cm,height=3cm]{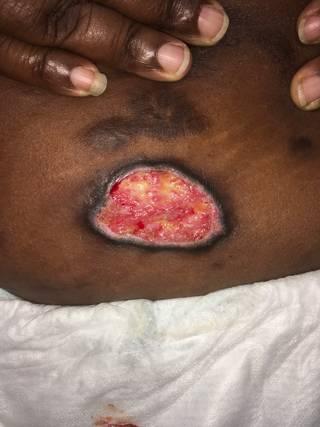}\quad
\includegraphics[width=2.7cm,height=3cm]{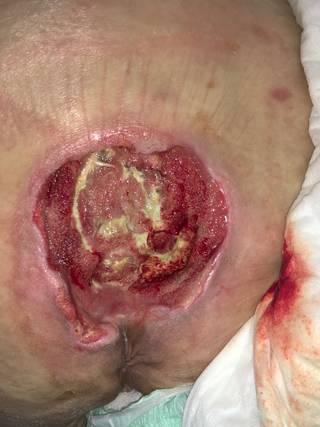}\quad
\includegraphics[width=2.7cm,height=3cm]{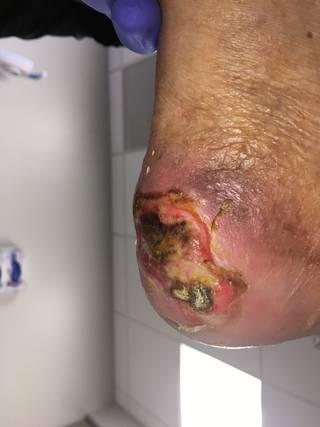}

\medskip

\includegraphics[width=2.7cm,height=3cm]{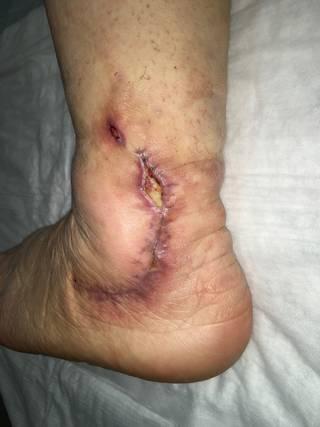}\quad
\includegraphics[width=2.7cm,height=3cm]{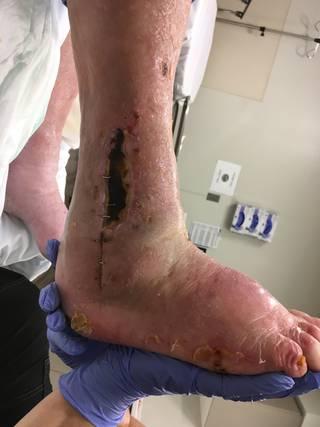}\quad
\includegraphics[width=2.7cm,height=3cm]{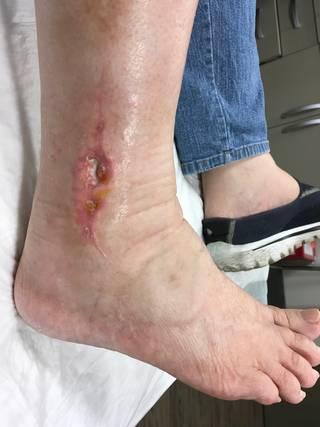}\quad
\includegraphics[width=2.7cm,height=3cm]{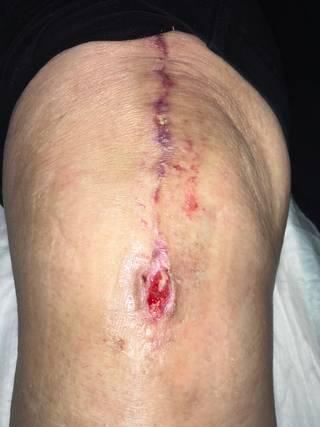}

\caption{Sample images from the AZH Wound and Vascular Center database. The rows from top to bottom display diabetic, venous, pressure, and surgical samples, respectively.}
\label{fig:AZHsamples}
\end{figure*}

In this research, we used a new wound image dataset collected over a two-year clinical period at the AZH Wound and Vascular Center in Milwaukee, Wisconsin. The dataset includes 400 wound images in jpg format and various sizes from four different wound types: venous, diabetic, pressure, and surgical (100 images per class). The images were captured using an iPad Pro (software version 13.4.1) and a Canon SX 620 HS digital camera and were labeled by a wound specialist from the AZH Wound and Vascular Center. The dataset can be accessed on GitHub with this link [https://github.com/uwm-bigdata/wound\_classification/blob/main/data/Dataset.rar].\\ Figure~\ref{fig:AZHsamples} shows some sample images from different classes of the dataset.

\subsection{Methods}
This section describes the method we used in this research and has been organized into two subsections: patch-wise and image-wise wound classification. The patch classifiers with the best performance will be used as a building block of the whole image classification model.

\subsubsection{Patch-wise classification}
\label{S:321}

\begin{figure*}[htp]
\centering
\includegraphics[width=2cm,height=2cm]{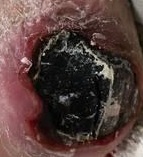}\quad
\includegraphics[width=2cm,height=2cm]{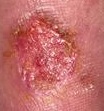}\quad
\includegraphics[width=2cm,height=2cm]{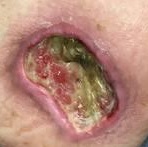}\quad
\includegraphics[width=2cm,height=2cm]{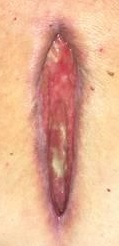}\quad
\includegraphics[width=2cm,height=2cm]{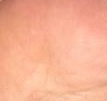}\quad
\includegraphics[width=2cm,height=2cm]{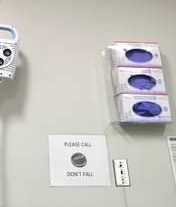}

\medskip

\includegraphics[width=2cm,height=2cm]{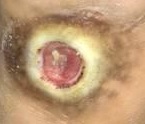}\quad
\includegraphics[width=2cm,height=2cm]{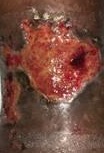}\quad
\includegraphics[width=2cm,height=2cm]{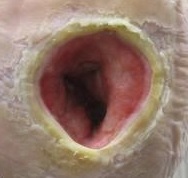}\quad
\includegraphics[width=2cm,height=2cm]{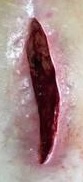}\quad
\includegraphics[width=2cm,height=2cm]{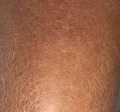}\quad
\includegraphics[width=2cm,height=2cm]{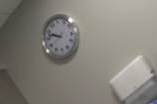}

\medskip

\includegraphics[width=2cm,height=2cm]{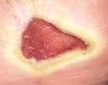}\quad
\includegraphics[width=2cm,height=2cm]{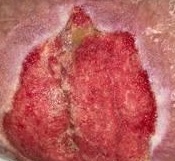}\quad
\includegraphics[width=2cm,height=2cm]{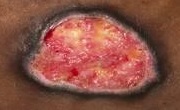}\quad
\includegraphics[width=2cm,height=2cm]{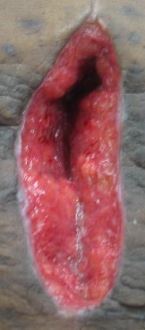}\quad
\includegraphics[width=2cm,height=2cm]{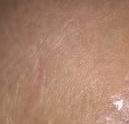}\quad
\includegraphics[width=2cm,height=2cm]{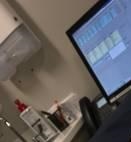}

\medskip

\includegraphics[width=2cm,height=2cm]{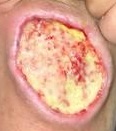}\quad
\includegraphics[width=2cm,height=2cm]{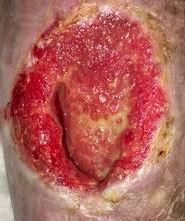}\quad
\includegraphics[width=2cm,height=2cm]{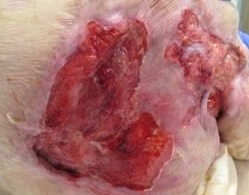}\quad
\includegraphics[width=2cm,height=2cm]{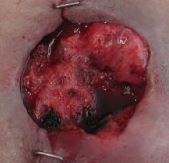}\quad
\includegraphics[width=2cm,height=2cm]{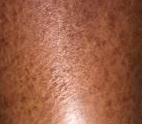}\quad
\includegraphics[width=2cm,height=2cm]{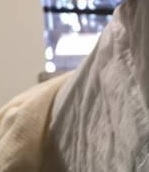}

\caption{Sample ROIs. The columns from left to right display diabetic, venous, pressure, surgical, normal skin, and background ROIs, respectively.}
\label{fig:ROI}
\end{figure*}

\paragraph{Pre-processing-ROI Extraction}~We selected 100 images for each wound type as the training samples and manually extracted 100 unique Regions of Interest (ROI) out of them per class, representing each of the six categories: diabetic, venous, pressure, surgical, background, and normal skin. The ROIs are rectangular and have different sizes. Figure~\ref{fig:ROI} displays some of the extracted ROIs from different classes of the dataset.

\paragraph{\textit{Pre-processing-Data splitting}}~After extracting the ROIs, we separated the samples by putting them into training (70\% samples), testing (15\% samples), and validation (15\% samples) sets.

\paragraph{\textit{Pre-processing-Patch generation}}~In this step, 17 patches were generated from each ROI, resulting in 1700 patches per class. The patches were extracted in a way that covers between 75\% to 85\% of the original ROI.  After this step, for each class we had 1190, 255, and 255 patches in the train, validation, and test sets, respectively.

\paragraph{\textit{Pre-processing-Data Augmentation}}~Augmentation of the training set samples was performed by generating 16 samples from each one using image transformation methods like rotating, flipping, cropping, and mirroring. Following augmentation, there were 19040 training samples in each class. \par

\begin{figure}[h!]

\centering\includegraphics[width=8.8cm,height=3.5cm]{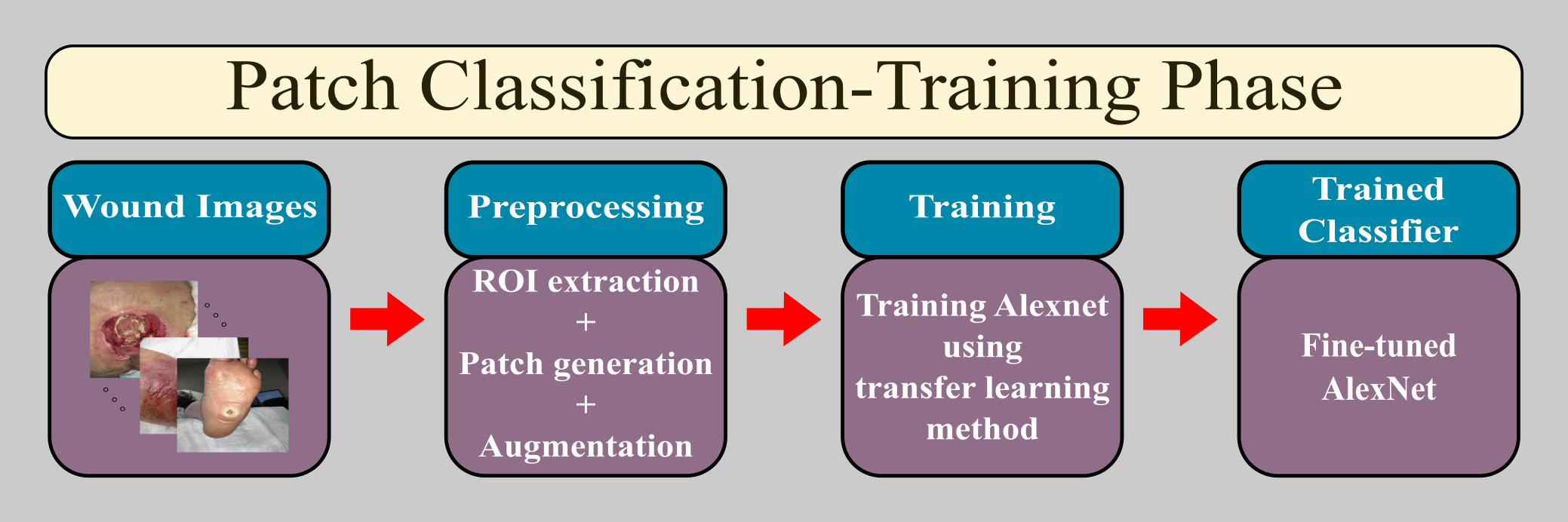}

\caption{Training process of the patch classifier.}
\label{fig:training}
\end{figure}


\paragraph{Training}\par
Following the pre-processing step, we trained a deep convolutional neural network using the training samples. After studying various networks, since our training set is small and the more modern CNNs need extensive number of samples and are computationally expensive, we selected AlexNet as the classifier. Due to its simplicity and effectiveness, AlexNet is still one of the most common deep networks used by researchers~\cite{litjens2017survey}. AlexNet is a deep CNN architecture proposed in 2012 and was the winner of ILSVRC in comparison to other traditional machine learning methods~\cite{krizhevsky2012imagenet,alom2018history}. This network has 8 layers and 60 million parameters. A modified version of AlexNet was utilized in this research to suit for our classification problems. We changed the fully connected layer in a way that its output size, matches the number of classes in our data. In addition, we used the transfer learning technique to increase the training accuracy while reducing the training time. It means that the AlexNet architecture was pre-trained on a massive dataset of general images called ImageNet and fine-tuned using our wound image patches. Figure~\ref{fig:training} shows the described steps for training the patch classifier.

\subsubsection{Image-wise classification using an ensemble classifier}
Several DCNN-based ensemble classification methods were proposed for medical or non-medical image classification tasks in the literature~\cite{chen2019deep,xia2019transferring,kassani2019classification}. Various strategies were used by the researchers to construct the ensemble classifier such as voting, concatenating, averaging, etc.~\cite{savelli2020multi,cha2019automated,hussain2020comprehensive, bermejo2020classification}. In all of these studies, the final conclusion was that the ensemble model outperformed the individual classifiers in performance. To this end, we designed an ensemble classifier in which the trained patch classifier described in Section~\ref{S:321} is used as a building block. In fact, the classification scores acquired from two classifiers (patch-wise and image-wise) are fed into an MLP classifier to obtain a better classification performance. Our hypothesis is that the proposed ensemble classifier will outperform each of the individual classifiers in terms of classification accuracy. Different components of the proposed classifier will be explained individually below.

\begin{figure}[h!]

\centering\includegraphics[width=8.5cm,height=3.5cm]{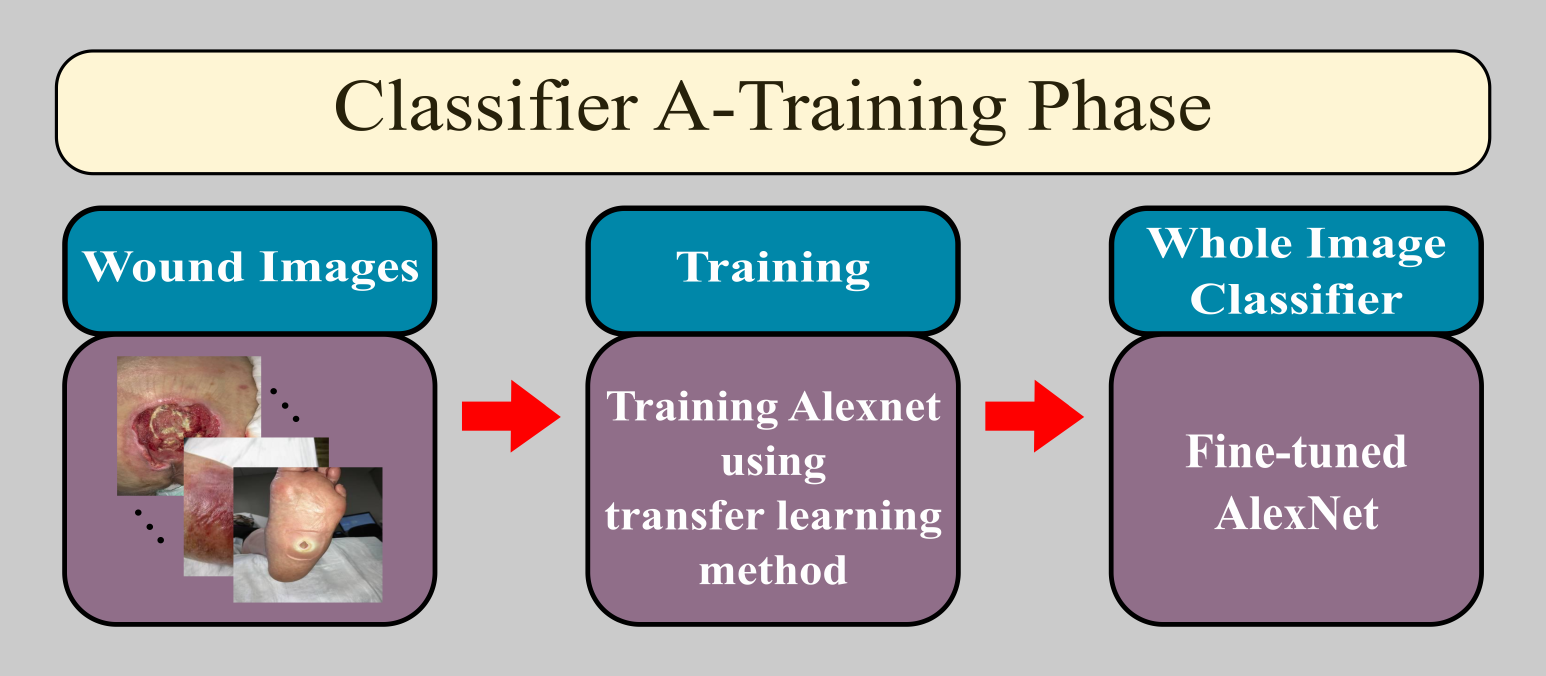}

\caption{Training process of Classifier A.}
\label{fig:classifierA}
\end{figure}

\paragraph{\textit{Classifier A - Whole image classifier}}~This classifier is a pre-trained AlexNet architecture that we fine-tuned using our own dataset. Figure~\ref{fig:classifierA} displays the training phase for this classifier.

\begin{figure*}[h!]

\centering\includegraphics[width=17cm,height=5.2cm]{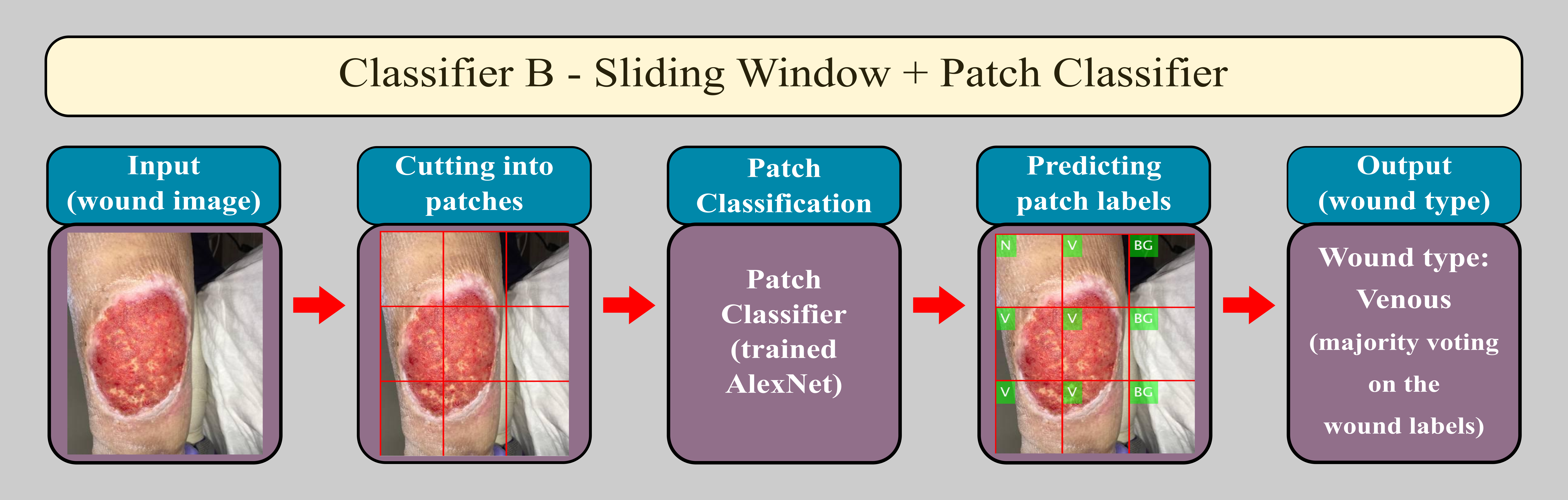}

\caption{Classifier B. The first step is extracting equal size patches out of the input image using the sliding window technique. Then the patch classifier is used for detecting the patch labels. The final step is majority voting for predicting the whole image label.}
\label{fig:classifierB}
\end{figure*}

\paragraph{\textit{Classifier B - Sliding window + Patch classifier}}~This classifier applies the sliding window technique on the input wound image to extract 9 patches of equal size along with patch classification step to predict their wound type. The wound type for the whole image will then be predicted by majority voting on the predicted label of the patches detected as wound. Figure~\ref{fig:classifierB} describes the entire process for this classifier.

\paragraph{\textit{Classification scores}}~Each of the classifiers A and B generates two classification scores for every input image. For example, for the surgical vs venous binary classification problem they output S and V scores which stand for classification scores of the surgical and venous labels, respectively. For the classifier B, these scores are calculated by averaging over S and V scores of all the patches detected as wounds by the patch classifier. In the end, for the binary classification case, we create a three-element feature vector including \( S_{A}, S_{B}\), and \(V_{B}\) in which the subscripts A and B show the related classifier. It is important to note that we did not include \(V_{A}\) in the feature vector, because of the correlation between \( S_{A}\) and \(V_{A}\). Finally, we feed the feature vector into the MLP classifier for the final classification task as described below.

\begin{figure*}[h!]

\centering\includegraphics[width=14.5cm,height=7.5cm]{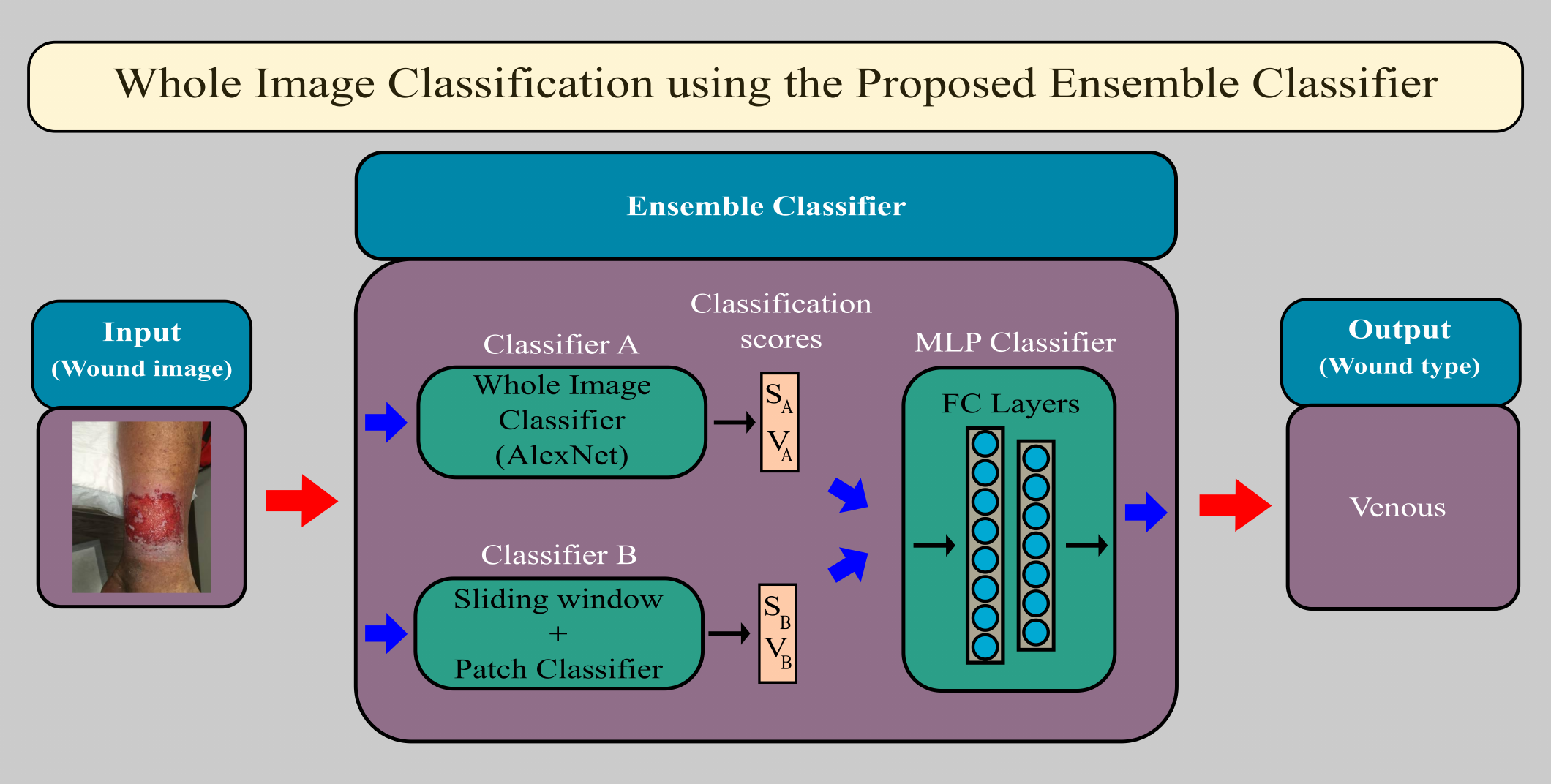}

\caption{Whole image classification process using our proposed ensemble classifier. The classifier accepts the wound image as the input and predicts the wound type as the output.}
\label{fig:ensemble}
\end{figure*}

\paragraph{\textit{MLP Classifier}}~The MLP classifier is a four-layer MLP with two hidden layers that have 8 and 7 neurons, respectively. The number of nodes in the input and output layers are determined based on the type of the classification problem. For the binary classification case, there are 3 and 2 nodes in these layers. The output of the MLP classifier is the wound type of the input image.
Figure~\ref{fig:ensemble} displays how the proposed ensemble classifier works in case of a binary classification problem. It is important to mention that the idea behind combining the Classifier A and B is to consider both patch level and whole image level information for classification.

\subsection{Performance metrics}
In this research we used accuracy, precision, recall, and F1-score metrics to investigate the performance of the classifiers. Equations~\ref{eq:acc} to~\ref{eq:score} show the related formulae for these evaluation metrics. In the binary classification problem, we used Area Under the ROC Curve (AUROC or AUC) metric as well. In these equations, TP, TN, FP and FN represent True Positive, True Negative, False Positive, and False Negative measures, respectively. More details about these equations and the related concepts can be found in~\cite{fawcett2006introduction}.

\begin{equation} \label{eq:acc}
\begin{split}
Accuracy = \frac{TP+TN}{TP+TN+FP+FN}
\end{split}
\end{equation}

\begin{equation} \label{eq:precision}
\begin{split}
Precision = \frac{TP}{TP+FP}
\end{split}
\end{equation}

\begin{equation} \label{eq:recall}
\begin{split}
Recall (TPR) = \frac{TP}{TP+FN}
\end{split}
\end{equation}

\begin{equation} \label{eq:score}
\begin{split}
F1-Score = 2 \times \frac{Precision\times Recall}{Precision+Recall}
\end{split}
\end{equation}



\section{Results}
\label{S:4}

This section presents the results obtained from the patch-wise and image-wise classification experiments. The classifiers were implemented in MATLAB R2019b and R2020a using an NVIDIA GEFORCE RTX 2080 Ti GPU with 11GB of memory. In the diagrams and tables presented in this section, the following abbreviations D, V, P, S, BG, and N represent the classes diabetic (D), venous (V), pressure (P), surgical (S), background (BG), and normal skin (N) (Table~\ref{tab:abb}). Several experiments were conducted to find the optimum structure for the deep networks. The optimum epoch number obtained was 20 and we used a learning rate value of 10e-6. Also, Adam was utilized as the optimization algorithm~\cite{kingma2014adam}. Further details for each experiment are provided below.  

\begin{table}[!t]
\centering
\caption{\\Class label abbreviations}
\label{tab:abb}
\begin{tabular}{|c|c|}
\hline
\textbf{Abbreviation} & \textbf{Description}  \\ \hline
          BG    &     Background         \\ \hline
            N  &      Normal skin       \\ \hline
          V     &     Venous           \\ \hline
           D    &     Diabetic          \\ \hline
            P   &      Pressure          \\ \hline
             S   &      Surgical          \\ \hline
\end{tabular}
\end{table}


\subsection{Patch classification}
\label{pclass}
\begin{figure*}[htp]
\centering
\includegraphics[width=7.3cm,height=7.3cm]{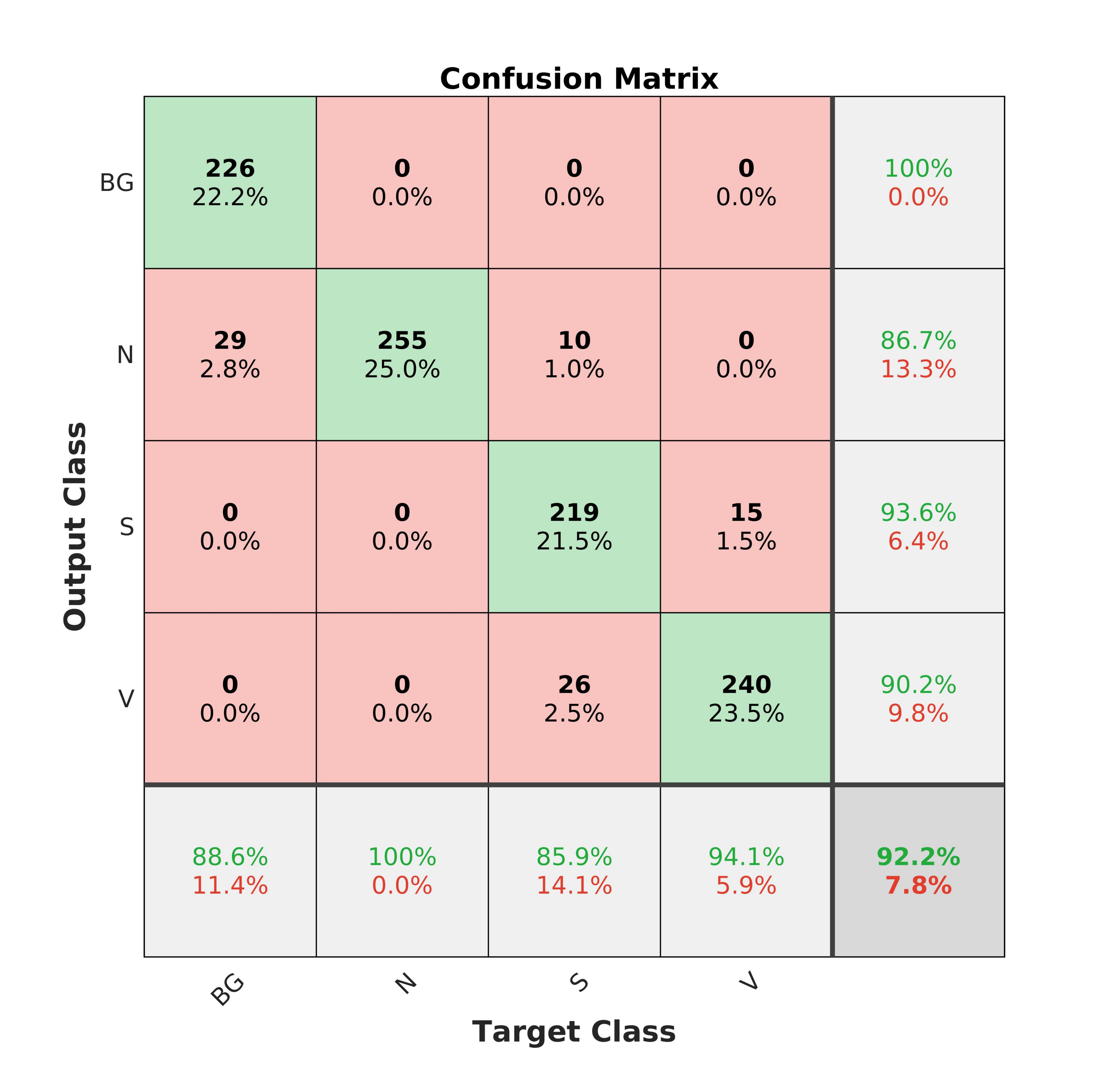}\quad
\includegraphics[width=7.3cm,height=7.3cm]{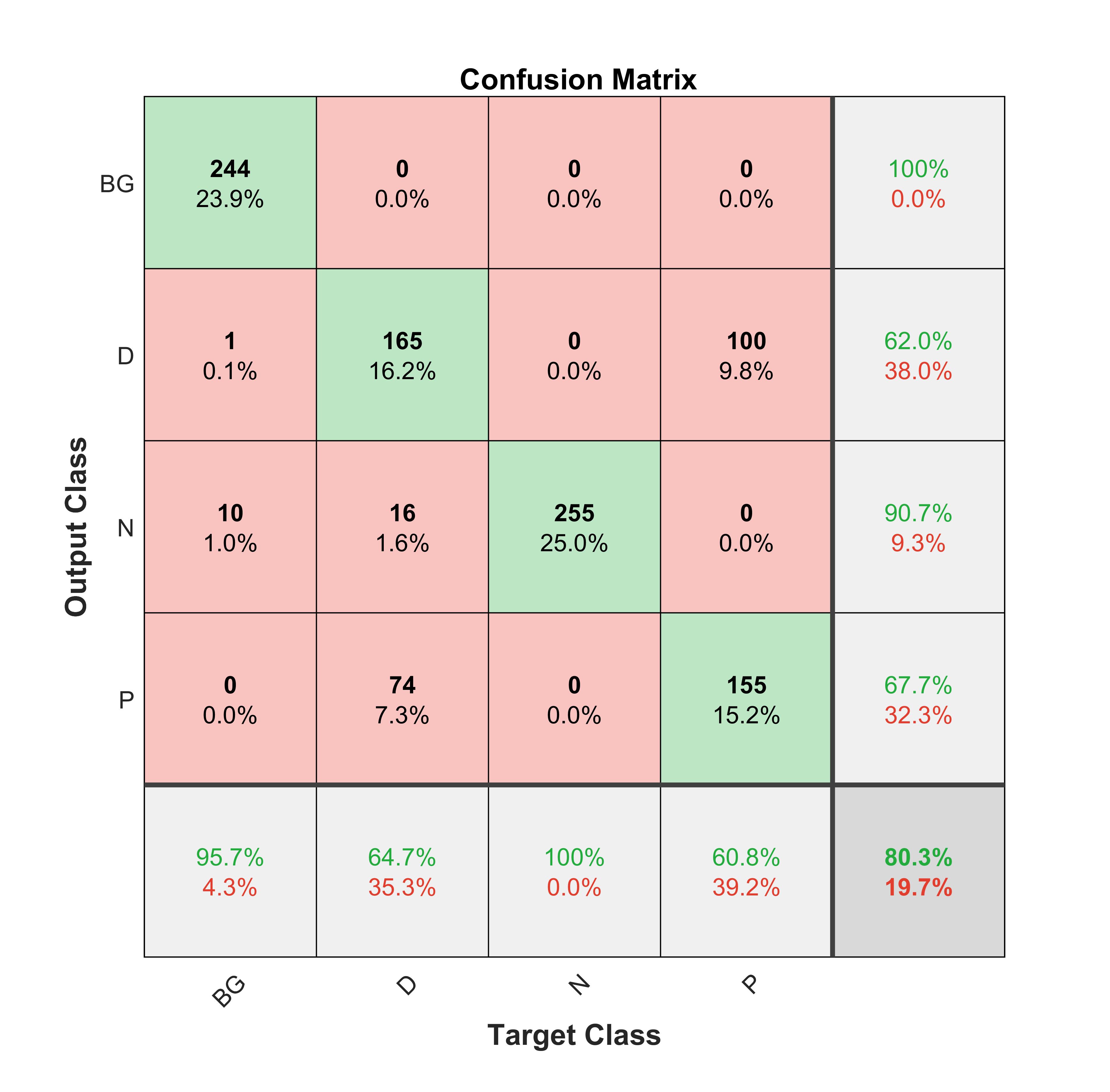}

\caption{Confusion matrices of the best (left) and worst (right) case in 4-class classification experiments.}
\label{fig:4classConf}
\end{figure*}

\begin{figure*}[htp]
\centering
\includegraphics[width=7.3cm,height=7.3cm]{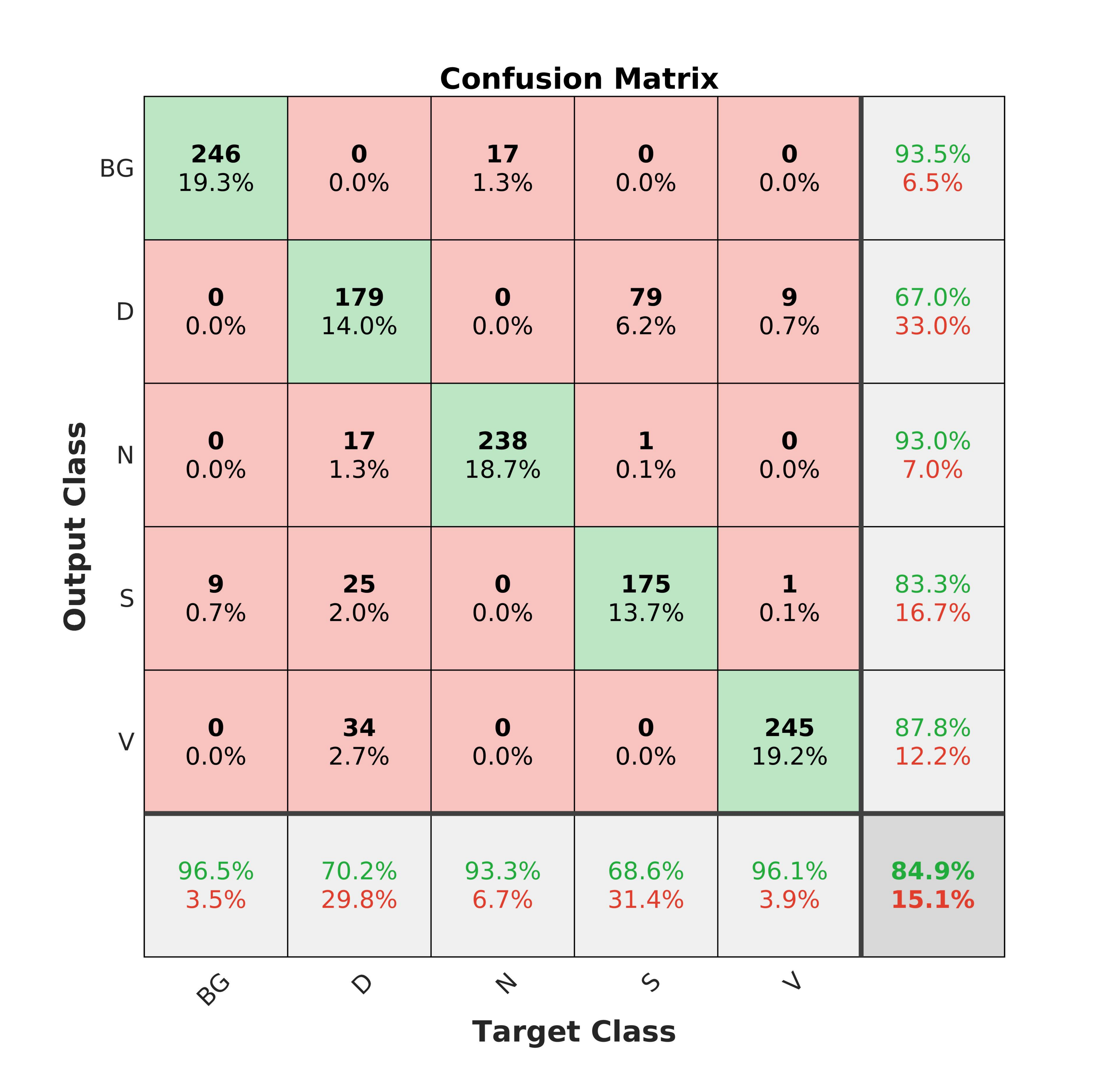}\quad
\includegraphics[width=7.3cm,height=7.3cm]{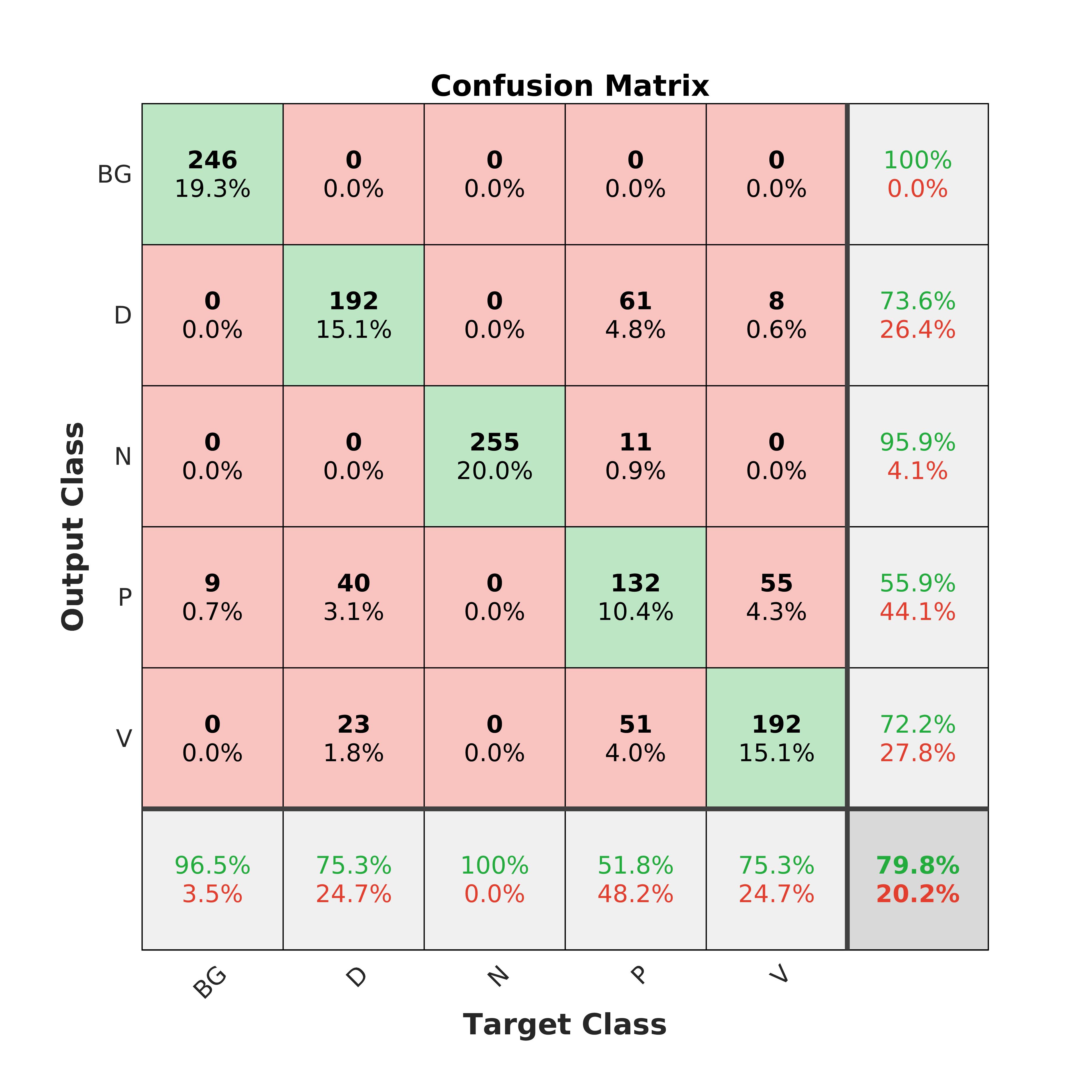}

\caption{Confusion matrices of the best (left) and worst (right) case in 5-class classification experiments.}
\label{fig:5classConf}
\end{figure*}

\begin{figure}[htp]
\centering
\includegraphics[width=7.3cm,height=7.3cm]{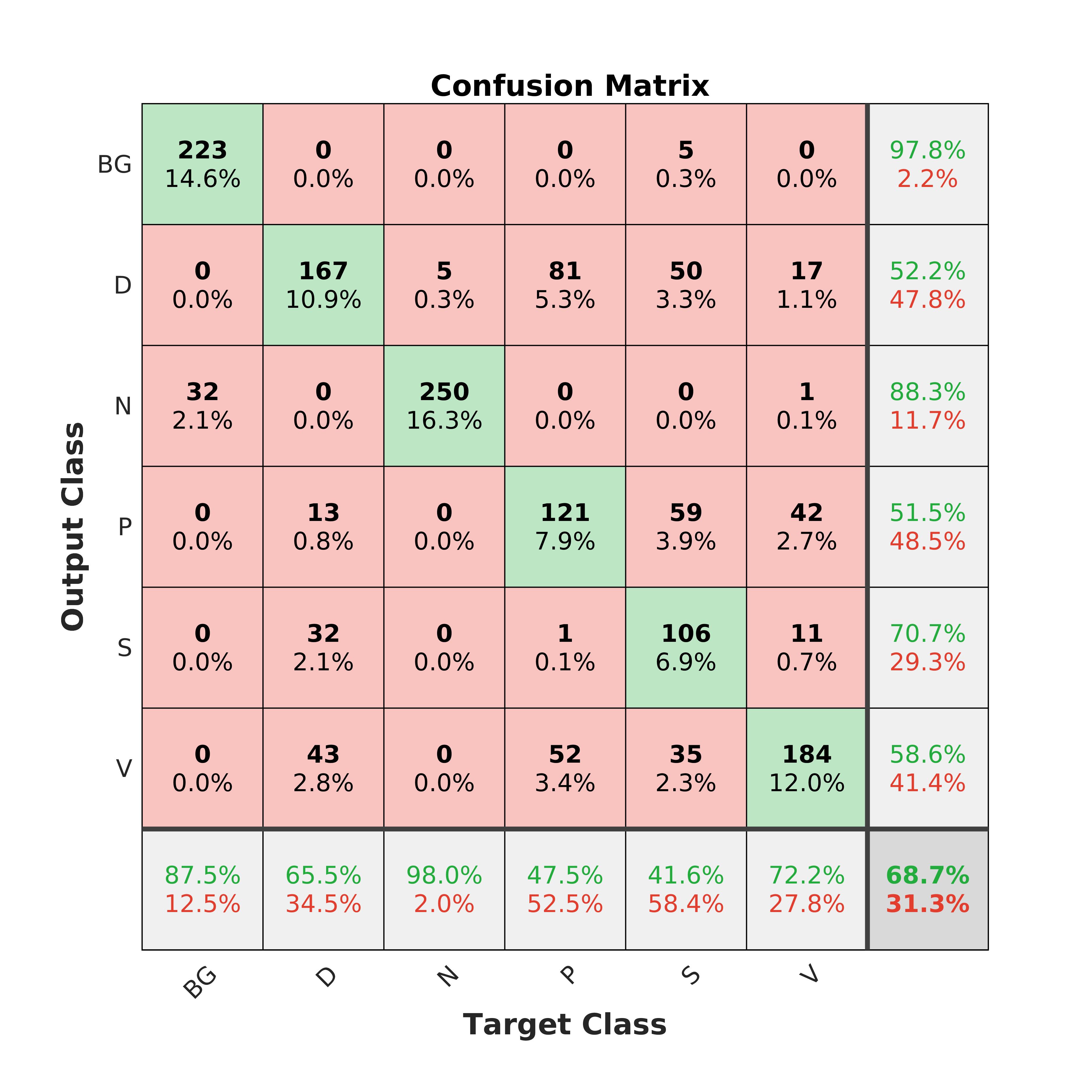}\quad

\caption{Confusion matrix for the 6-class classification experiment.}
\label{fig:6classConf}
\end{figure}

To evaluate the patch classifier's performance for patch-wise classification, we used 255 test patches per class. For 4-class classification experiments, the goal was to classify the wound patches into one of the four classes: BG, N, and two wound labels. In the 5-class classification problem, we had three wound labels as well as the BG and N classes. The last group of the patch-wise classification experiments is related to the 6-class classification case in which the wound patches are classified into one of the six classes diabetic, venous, pressure, surgical, BG, and N. Table~\ref{tab:4cwoaug} shows the test accuracy values for all the experiments mentioned above. Figures~\ref{fig:4classConf} to \ref{fig:6classConf} display some sample confusion matrices for patch-wise classification experiments. It should be noted that we performed and compared all the experiments with and without data augmentation. As data augmentation always resulted in better results, we only show our experiments with data augmentation.

\begin{table}[h!]
\caption{\\Patch-wise classification results.}
\label{tab:4cwoaug}
\centering
\begin{tabular}{|c|c|c|}
\hline
\textbf{Num of Classes} & \textbf{Classes} & \textbf{Test accuracy (\%)} \\ \hline
        &  BGNVD            &       89.41       \\ \cline{2-3}
         &   BGNVP          &      86.57        \\ \cline{2-3}
    4-class    &  BGNVS           &       \textbf{92.20}      \\ \cline{2-3}
         &  BGNDP          &        80.29      \\ \cline{2-3}
          &  BGNDS         &       90.98      \\ \cline{2-3}
          &   BGNPS        &       84.12       \\ \hline
        &  BGNDVP         &      79.76       \\ \cline{2-3}
   5-class    &     BGNDVS       &       \textbf{84.94}       \\ \cline{2-3}
        &  BGNDPS         &      81.49      \\ \cline{2-3}
        &   BGNVPS         &       83.53       \\ \hline
   6-class    &   BGNDVPS          &       \textbf{68.69}      \\ \hline
\end{tabular}
\end{table}

\subsection{Whole image classification}
To assess our proposed ensemble classifier's efficiency, we performed two types of experiments: binary classification and 3-class classification. With the patch classification results presented in subsection~\ref{pclass}, surgical vs venous and surgical vs venous vs diabetic classifiers were selected which showed the best binary and 3-class classification outcomes. For the rest of the manuscript, we name the whole image classifier (trained AlexNet on the whole wound images) as Classifier A, and the patch classifier (trained AlexNet on the wound patches) as Classifier B. We also obtained 138 extra wound images from three classes surgical (28 samples), diabetic (54 samples), and venous (56 samples) from the AZH wound and vascular center to be used as the test images. Table~\ref{tab:wholebinary} shows the binary classification results obtained from applying the Classifier A, Classifier B, and the proposed ensemble classifier on the test set which included 84 wound images. Also, the 3-class classification results for the three classifiers have been provided in Table~\ref{tab:3classtest}.
The superiority of our proposed ensemble classifier over the other two classifiers can be detected from these results for both binary and 3-class classification problems.

\begin{table}[h!]

\caption{\\Whole image binary classification (surgical vs venous) results obtained from applying the classifiers on the test set images.}
\label{tab:wholebinary}
\centering
\begin{tabular}{|c|c|c|c|c|}
\hline



\multirow{3}{*}{\textbf{Classifier} } & & &  &    \\
    & \textbf{Accuracy} & \textbf{Precision} & \textbf{Recall} & \textbf{F1-Score}   \\
     & \textbf{(\%)} & \textbf{(\%)} &\textbf{(\%)}   &  \textbf{(\%)} \\ \hline

 A                                                               & 83.3                   & 76.9                    & 71.4                 & 74.04              \\ \hline
 B                                                               & 82.1                   & 71                      & 78.6                 & 74.60              \\ \hline
\begin{tabular}[c]{@{}c@{}}\textbf{Our}\\\textbf{ensemble} \\\textbf{classifier}\end{tabular} & \textbf{96.4}                   & \textbf{93.1}                    & \textbf{96.4}                 & \textbf{94.72}              \\ \hline
\end{tabular}

\end{table}

 \begin{table}[h!]
 \small
 \setlength{\arrayrulewidth}{0.25mm}
\caption{\\Whole image 3-class classification results obtained from applying the classifiers on the test set images.}
\label{tab:3classtest}
\centering
\begin{tabular}{|c|c|c|c|c|c|}\hline
\multirow{3}{*}{ \textbf{Classifier}} &  &  &  &  &  \\
    & \textbf{Class} & \textbf{Precision} & \textbf{Recall} & \textbf{F1-score} & \textbf{ACC}  \\
    &  & \textbf{(\%)} & \textbf{(\%)} & \textbf{(\%)}  & \textbf{(\%)}  \\
\noalign{
\hrule height 1.5pt
} 
\multirow{3}{*}{ \textbf{A}} & D & 88 & 81.5 & 84.62 &  \\\cline{2-5}
    & S & 67.9 & 67.9 & 67.9 & 83.3  \\\cline{2-5}
    & V & 86.7 & 92.9 & 89.69 &  \\ \noalign{
\hrule height 1.5pt
} 
    \multirow{3}{*}{\textbf{B}} & D & 70.9 & 72.2 & 71.54 &   \\\cline{2-5}
    & S & 42.1 & 28.6 & 34.06 & 67.4  \\\cline{2-5}
    & V & 71.9 & 82.1 & 76.66 &   \\
    \noalign{
\hrule height 1.5pt
} 
  \multirow{3}{*}{  \begin{tabular}[c]{@{}c@{}} \textbf{Our} \\ \textbf{ensemble} \\ \textbf{classifier} \\ 
  \end{tabular}} & D & 86.2 & 92.6 & 89.28 &   \\\cline{2-5}
    & S & 81.5 & 78.6 & 80.02 & 89.1  \\\cline{2-5}
    & V & 96.2 & 91.1 & 93.58 &  \\
    \noalign{
\hrule height 1.5pt
} 
\end{tabular}
\end{table}
 

\section{Discussion}
\label{S:5}
Acute and chronic wounds are a challenge and burden to healthcare systems in all countries. The wound diagnosis and treatment process can be facilitated using an efficient classification method. ML and DL have a good potential to be used as powerful algorithms for wound image analysis tasks such as classification. Prior works in the literature mainly dealt with binary classification or studied only specific types of wounds like diabetic ulcers. Additionally, the major part of the previous studies classified extracted ROIs or wound patches, rather than the whole wound images. Also, most of them had difficulties accessing valid, reliable, and high-quality images as some of them collected their data from the web. For these reasons, we proposed an end-to-end novel ensemble deep learning-based classification method for classifying the chronic wounds into multi categories based on their type.\par In the patch classification, as we expected, by increasing the number of classes from four to five and six, the classification accuracy decreases. The justification for this phenomenon is that increasing the number of classes accordingly increase the number of network parameters, which would make it more challenging for the deep architecture to train all the parameters to the same standard as before, using the same number of training samples. Another interesting observation is that in the 4-class and 5-class classification experiments, the lowest classification accuracy is related to the diabetic and pressure wounds. It shows that these two wound types are very similar in appearance. The confusion matrices confirm this fact by showing that many diabetic wounds had been classified into pressure class and vice versa. 
 Also, we observed in these experiments, the highest classification accuracy is related to the surgical wounds. It can be concluded that surgical wounds are the most distinguishable wound type from others. This could be related to the shapes of the surgical wounds which are usually more elongated and therefore more distinguishable among other wound types. We see that, in most of the experiments, the pressure wound is the most challenging wound type to classify. By looking closely at the confusion matrices, we find that the low recall value for this wound type often comes from misclassifying the wound into the venous or diabetic instead of the pressure class. 
 Another observation is that the venous wounds typically show the highest recall values among all of the wound types. This phenomenon could be related to having samples from a wider variability and consequently better training of the classifier for this wound type. We expect  that increasing the number of samples for dataset categories would improve the recall value for all the wound types. In all of the patch classification experiments, the background and normal skin classes have the highest recall values. This is important because in our proposed ensemble classifier we need to have a patch classifier with the ability to distinguish background and normal skin parts from the wounded tissue with a high accuracy value.\par About the image-wise classification experiments we see that for the binary case, both Classifier A and B generate almost similar results while our proposed ensemble classifier showed an accuracy value of 96.4\% which is 13.1\% higher than the Classifier A and 14.3\% higher than the Classifier B. Also, for the 3-class classification case, the accuracy of the ensemble classifier is higher than the other two classifiers. This last observation is very interesting because the Classifier B displays a low classification performance specifically for the surgical wounds, but after combining with the Classifier A, its accuracy value is improved by 5.8\%. 
 
 \subsection{Robustness experiments}
\begin{figure}[h!]

\centering\includegraphics[width=0.9\linewidth]{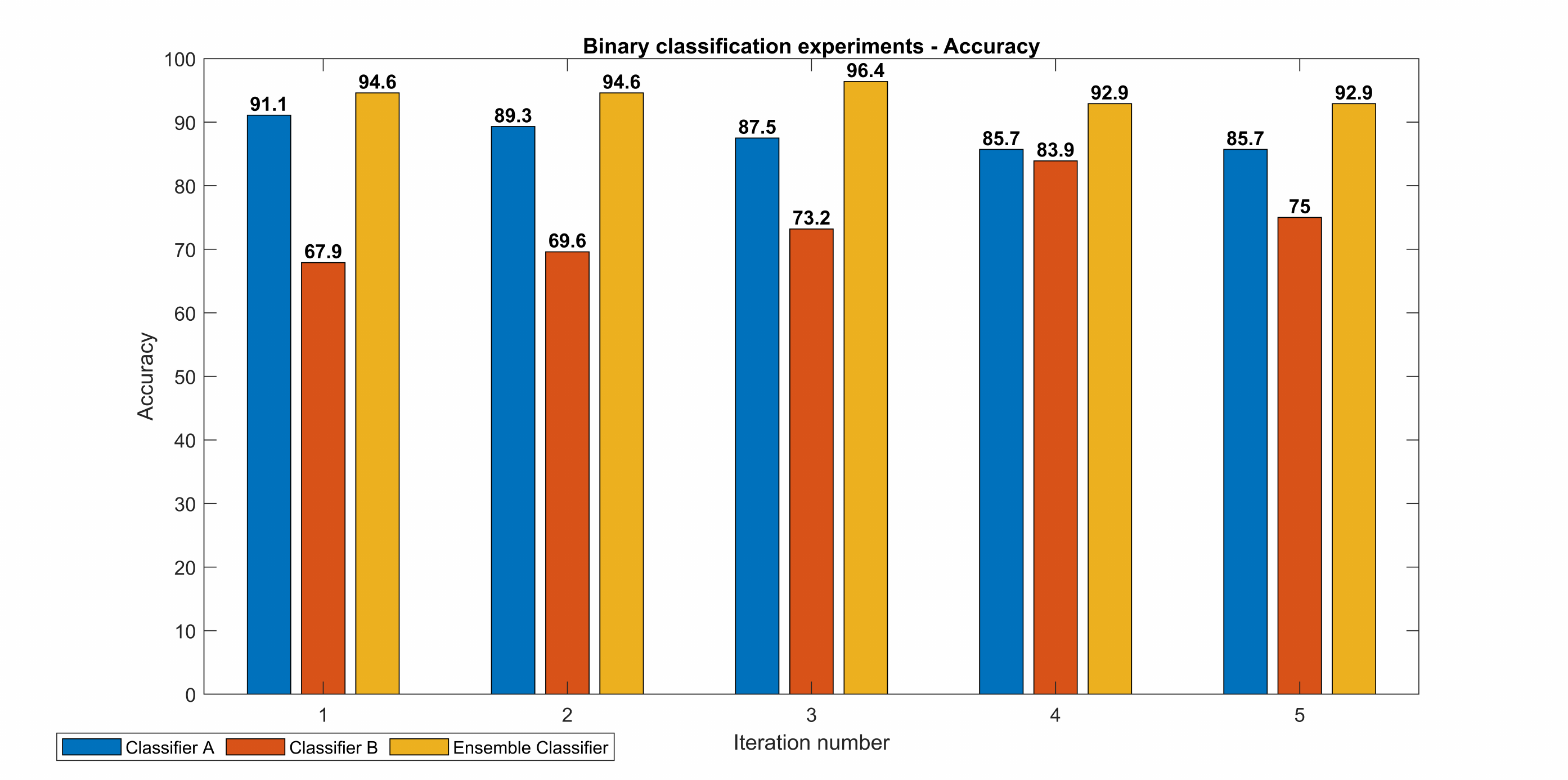}

\caption{Accuracy values obtained from 5-fold cross-validation for the binary classification problem.}
\label{fig:binaryACC}
\end{figure}

\begin{figure}[h!]

\centering
\includegraphics[width=1.0\linewidth]{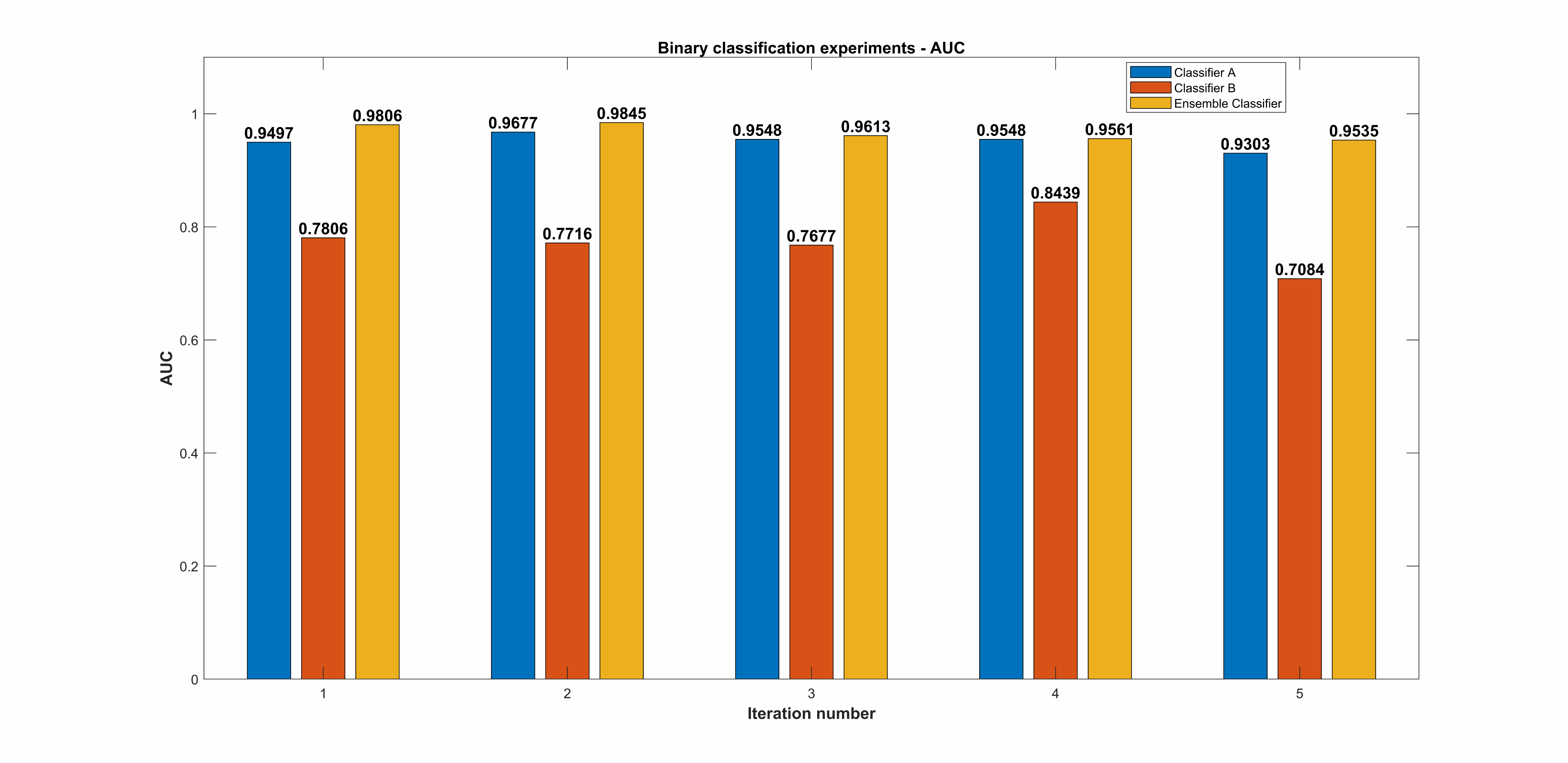}

\caption{AUC values obtained from 5-fold cross-validation for the binary classification problem.}
\label{fig:binaryAUC}
\end{figure}

\begin{figure*}[htp]
\centering

    \begin{subfigure}[b]{0.3\textwidth}
        \includegraphics[width=4.85cm,height=4.85cm]{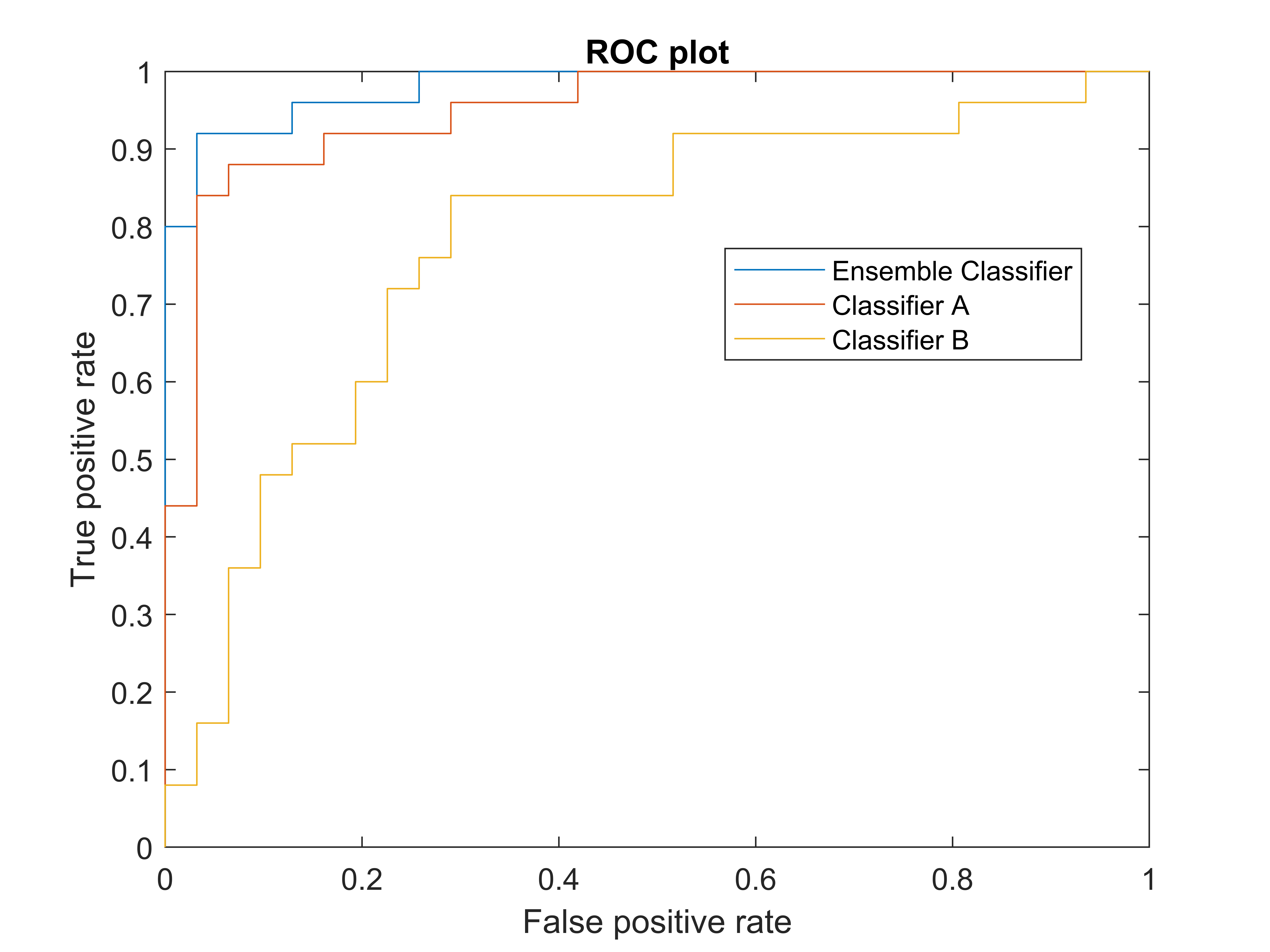}
        \caption{Round 1}
        \label{fig:gull}
    \end{subfigure}
    ~ 
    \begin{subfigure}[b]{0.3\textwidth}
        \includegraphics[width=4.85cm,height=4.85cm]{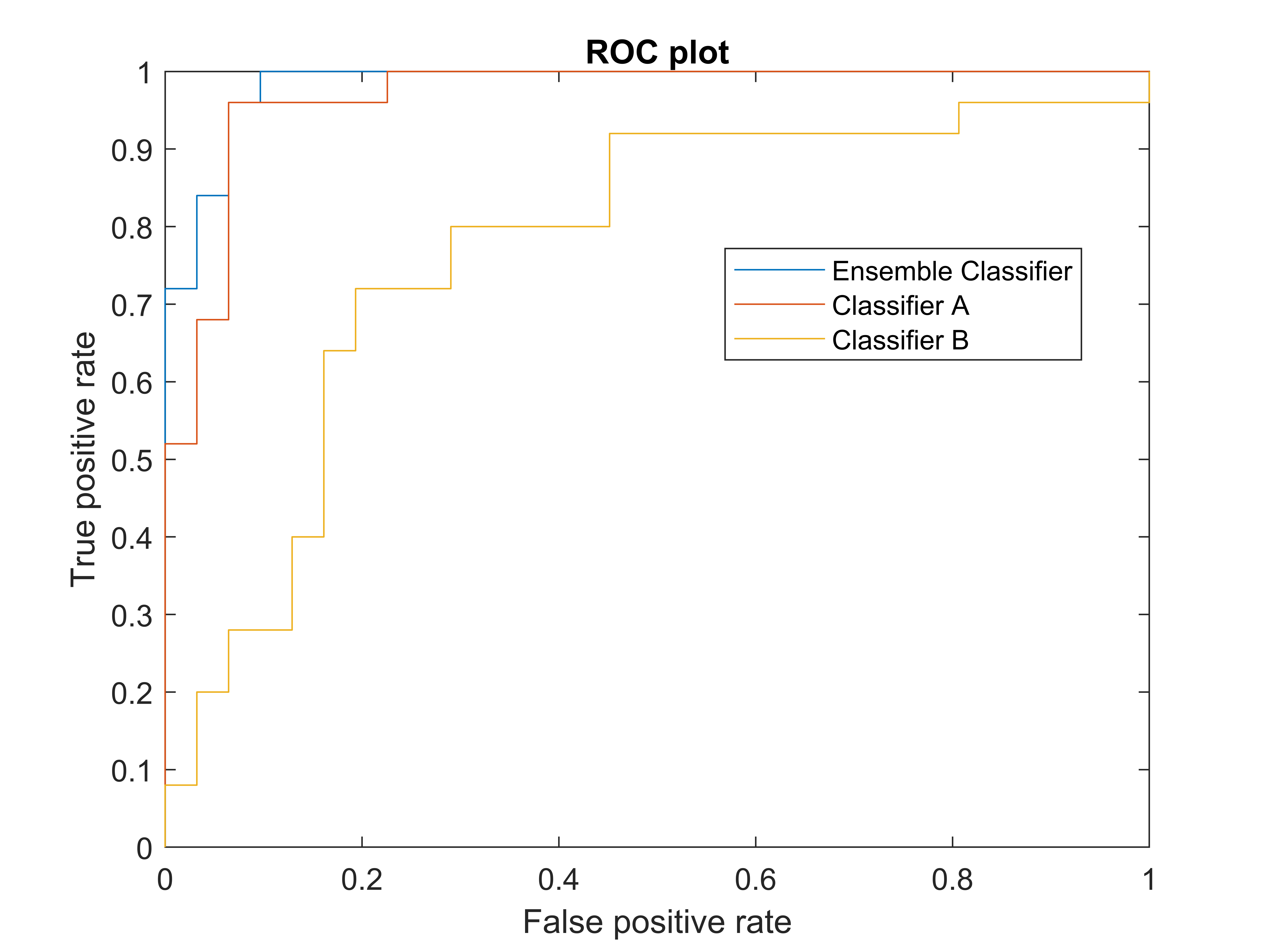}\quad
        \caption{Round 2}
        \label{fig:tiger}
    \end{subfigure}
    ~ 
    \begin{subfigure}[b]{0.3\textwidth}
        \includegraphics[width=4.85cm,height=4.85cm]{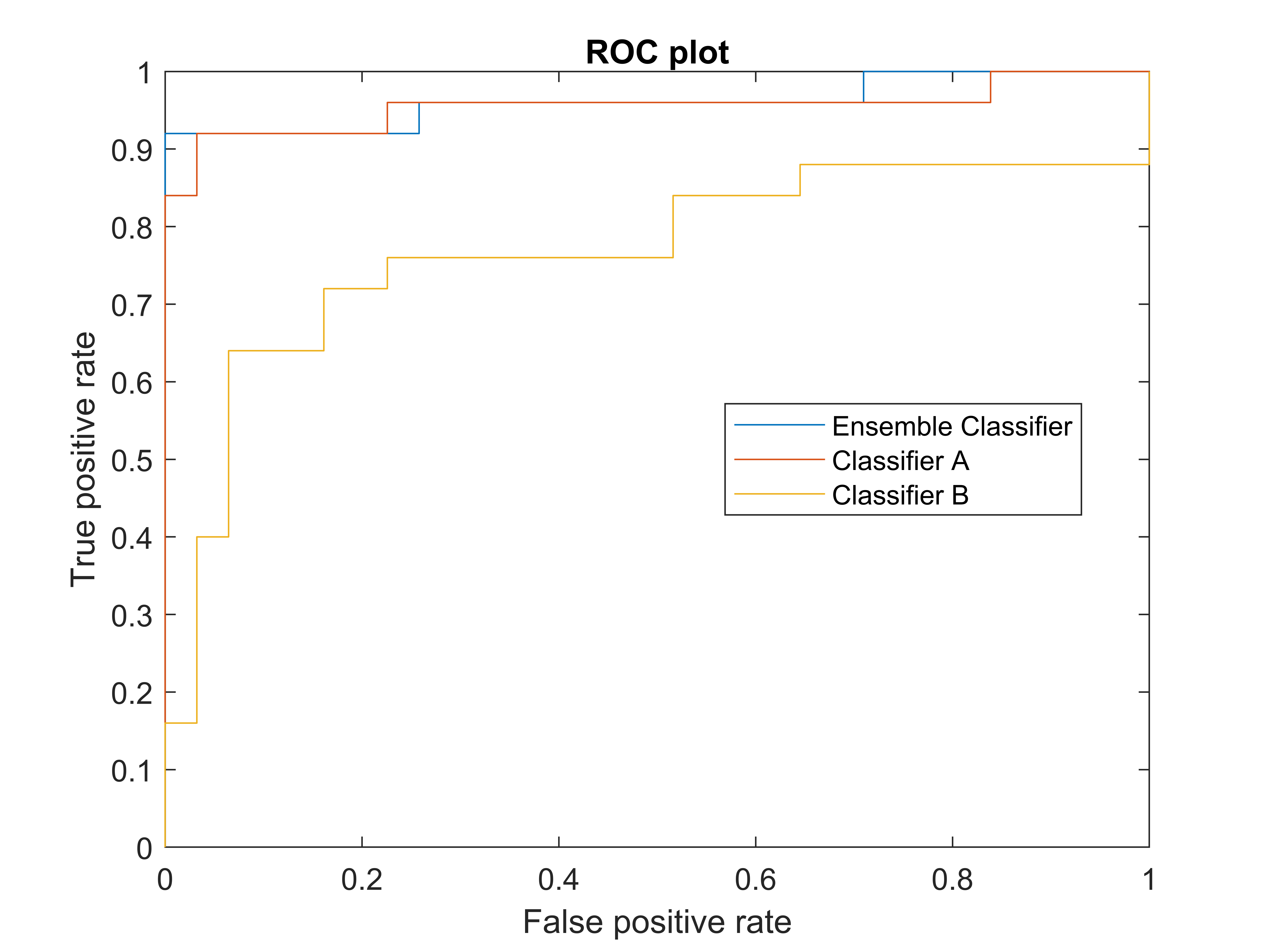}
        \caption{Round 3}
        \label{fig:mouse}
    \end{subfigure}
    ~ 
    \begin{subfigure}[b]{0.3\textwidth}
        \includegraphics[width=4.85cm,height=4.85cm]{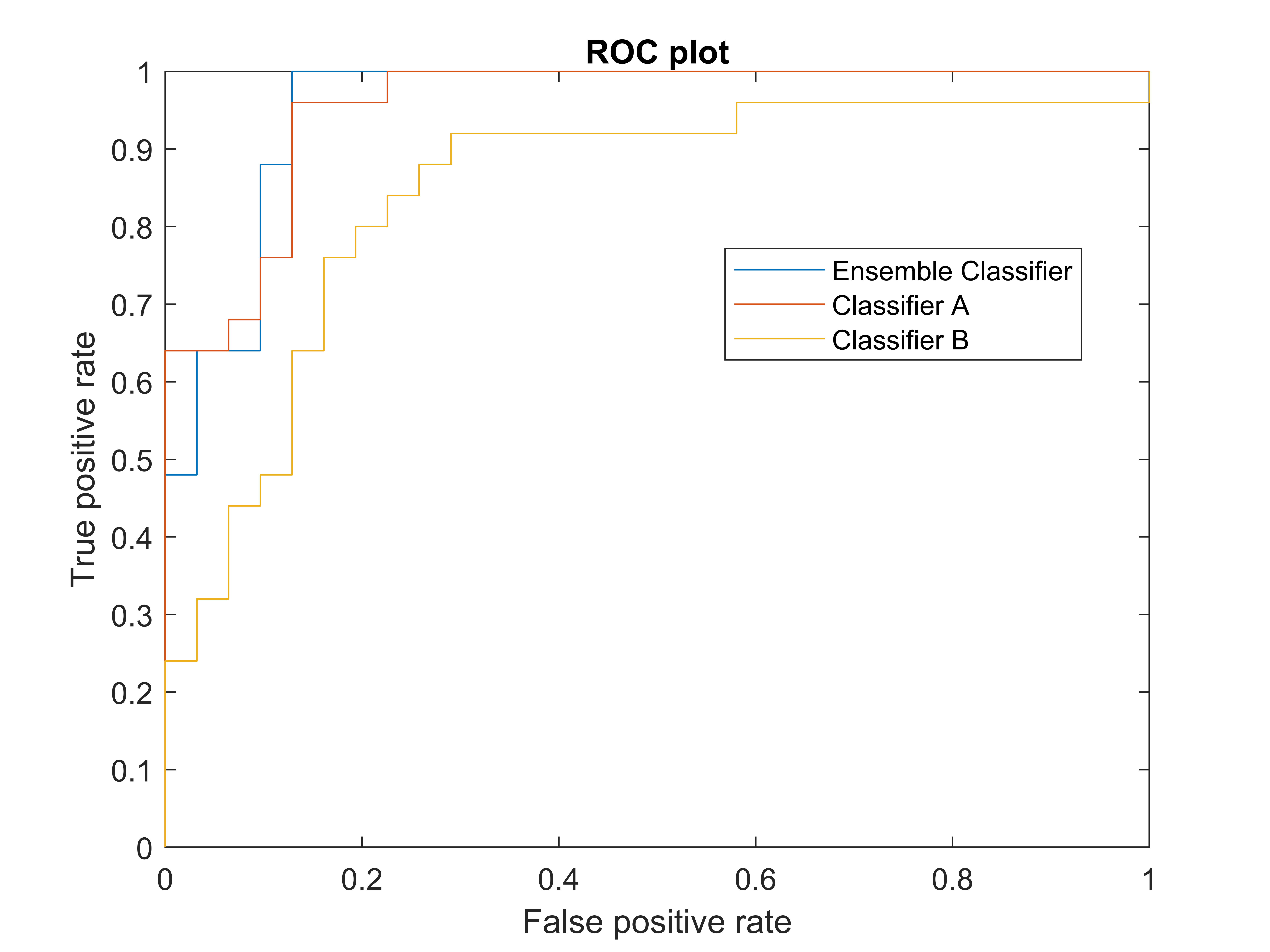}
        \caption{Round 4}
        \label{fig:mouse}
    \end{subfigure}
    ~ 
    \begin{subfigure}[b]{0.3\textwidth}
        \includegraphics[width=4.85cm,height=4.85cm]{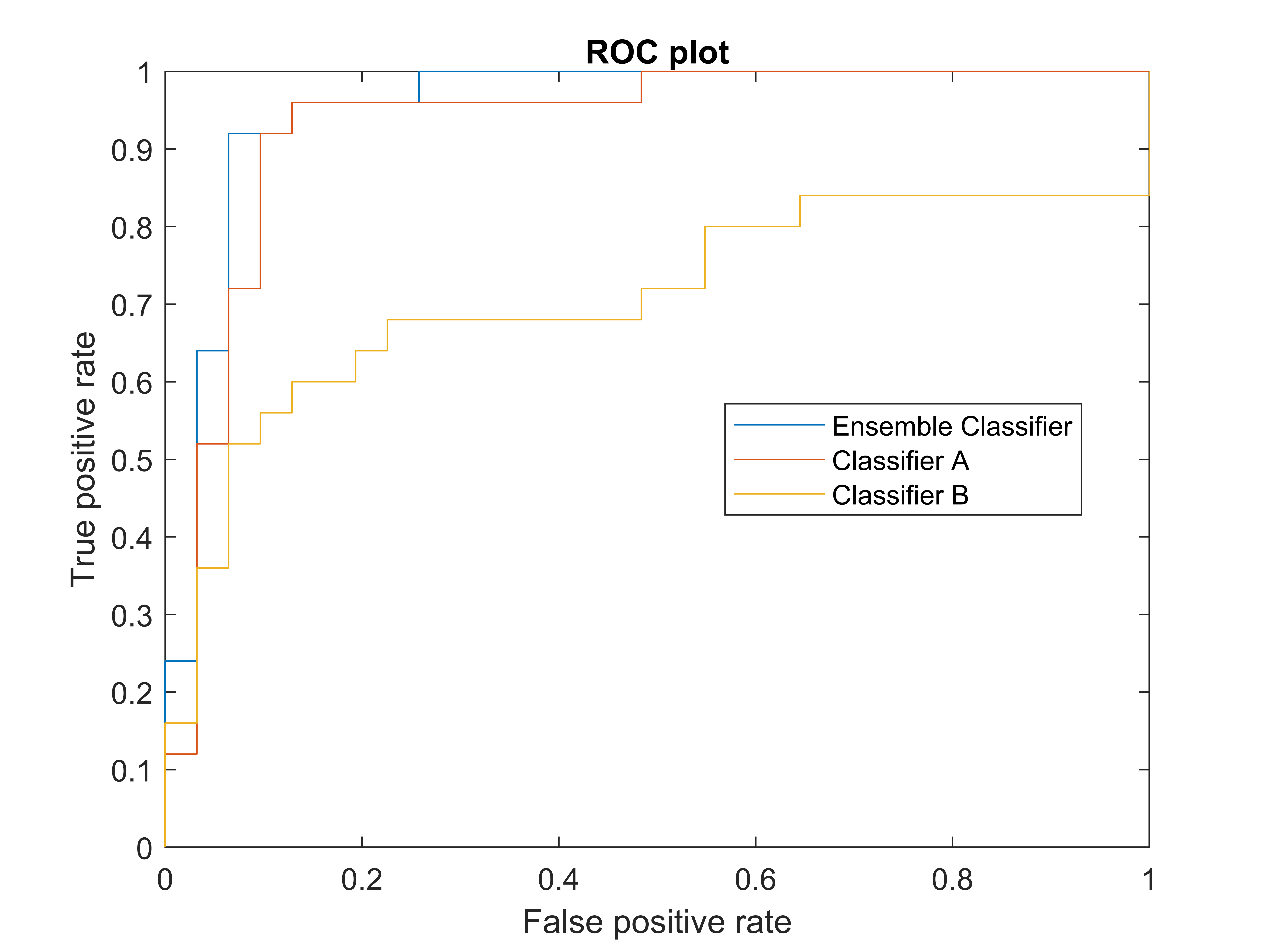}
        \caption{Round 5}
        \label{fig:mouse}
    \end{subfigure}

\caption{ROC plots obtained from 5-fold cross-validation experiments.}
\label{fig:roc}
\end{figure*}

\begin{figure}[h!]

\centering\includegraphics[width=0.9\linewidth]{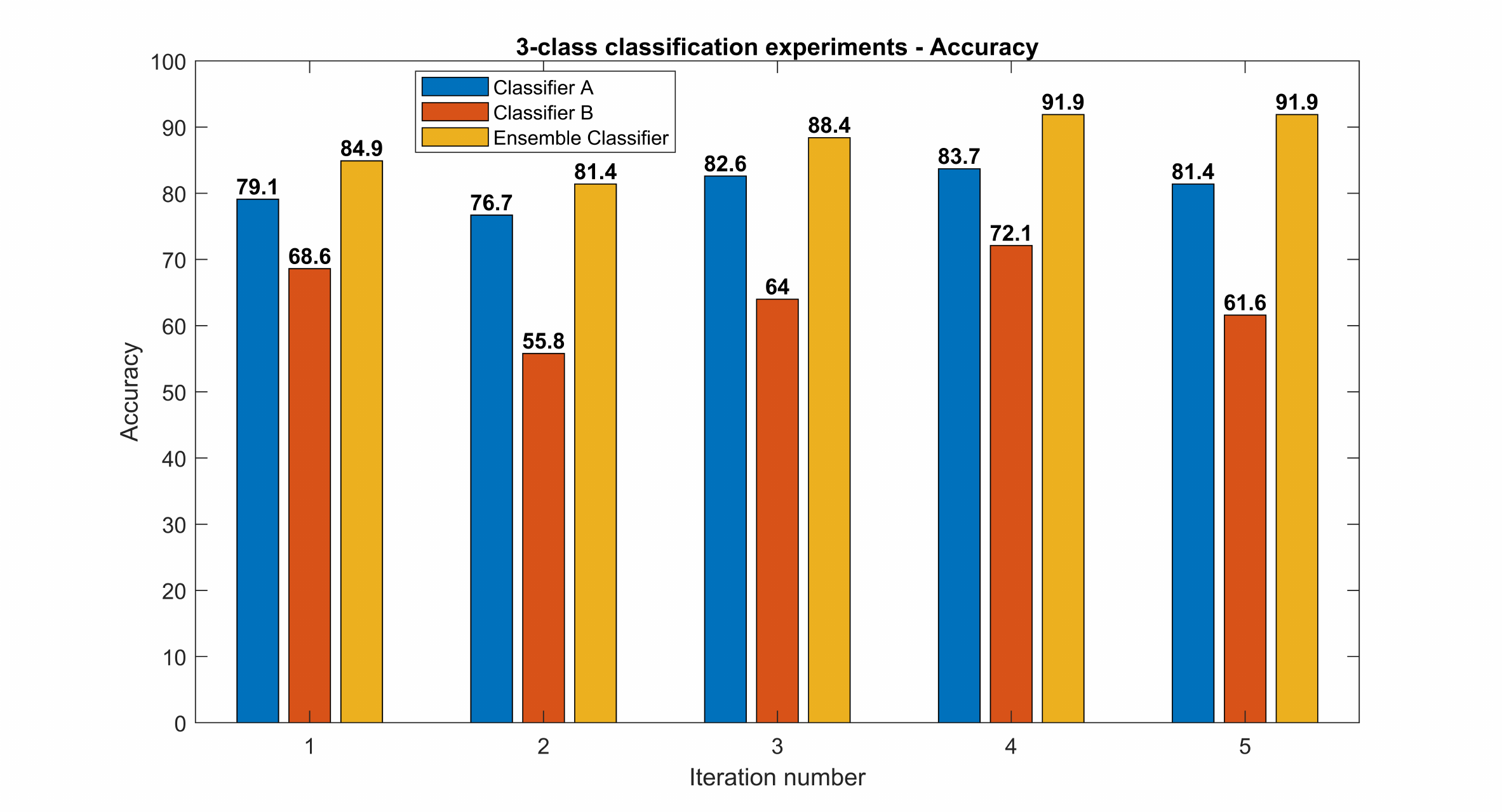}

\caption{Accuracy values obtained from 5-fold cross-validation for the 3-class classification problem.}
\label{fig:3classACC}
\end{figure}

 To investigate the robustness of our proposed classifier, we used 5-fold cross-validation as a standard evaluation method. For these experiments, we added the test set (84 images for binary and 138 images for 3-class classification experiment) to our training set (200 images for binary and 300 images for 3-class classification case). Thereafter, for each of the three classes diabetic, surgical, and venous, we selected 20\% of the samples randomly as the test set and the rest as the training samples. We trained our classifier using the training samples and tested it on the test set and repeated this strategy for 5 iterations. Tables~\ref{tab:accbinary} to~\ref{tab:F1} display the classification accuracy, AUC, precision, recall, and F1-score values obtained for the binary classification problem. Figures~\ref{fig:binaryACC} and \ref{fig:binaryAUC} compares the classifiers in accuracy and AUC metrics. The ROC plots presented in Figure~\ref{fig:roc}. Also, 3-class classification results have been summarized in Table~\ref{tab:acc3class} to \ref{tab:fscore3class} and Figure~\ref{fig:3classACC}. In all tables, R1 to R5 display the round number of the experiments. 
\begin{table}[h!]
\caption{\\Whole image binary classification (S vs V) accuracy percentages obtained from 5-fold cross-validation.}
\label{tab:accbinary}
\centering
\begin{tabular}{|c|c|c|c|c|c|}
\hline
\textbf{Classifier}                                                        & \textbf{R 1} & \textbf{R 2} & \textbf{R 3} & \textbf{R 4} & \textbf{R 5} \\ \hline
 A                                                               & 91.1                   & 89.3                    & 87.5                 & 85.7 & 85.7       \\ \hline
 B                                                               & 67.9                  & 69.6     & 73.2 & 83.9 &75              \\ \hline
\begin{tabular}[c]{@{}c@{}}\textbf{Our ensemble}\\ \textbf{classifier}\end{tabular} & \textbf{94.6}                   & \textbf{94.6}                    & \textbf{96.4}  & \textbf{92.9}& \textbf{92.9}           \\ \hline
\end{tabular}
\end{table}


\begin{table}[h!]
\caption{\\Whole image binary classification (S vs V) AUC values obtained from 5-fold cross-validation.}
\label{tab:aucbinary}
\centering
\begin{tabular}{|c|c|c|c|c|c|}
\hline
\textbf{Classifier}                                                        & \textbf{R 1} & \textbf{R 2} & \textbf{R 3} & \textbf{R 4} & \textbf{R 5} \\ \hline
 A                  & 0.9497                   & 0.9677                    & 0.9548                 & 0.9548 & 0.9303       \\ \hline
 B                                                               & 0.7806      & 0.7716     & 0.7677 & 0.8439 & 0.7084              \\ \hline
\begin{tabular}[c]{@{}c@{}}\textbf{Our ensemble}\\ \textbf{classifier}\end{tabular} & \textbf{0.9806}                   & \textbf{0.9845}                    & \textbf{0.9613}  & \textbf{0.9561}& \textbf{0.9535}           \\ \hline
\end{tabular}
\end{table}


\begin{table}[h!]
\caption{\\Whole image binary classification (S vs V) Precision percentages obtained from 5-fold cross-validation.}
\label{tab:precision}
\centering
\begin{tabular}{|c|c|c|c|c|c|}
\hline
\textbf{Classifier}                                                        & \textbf{R 1} & \textbf{R 2} & \textbf{R 3} & \textbf{R 4} & \textbf{R 5} \\ \hline
 A                                            & 91.7      & 91.3                    & 100   & 84 & 87       \\ \hline
 B                                          & 65.2      & 68.2     & 77.8 & 83.3 &76.2      \\ \hline
\begin{tabular}[c]{@{}c@{}}\textbf{Our ensemble}\\  \textbf{classifier}\end{tabular} & \textbf{95.8}                   & \textbf{89.3}                    & \textbf{100}  & \textbf{86.2}& \textbf{92}           \\ \hline
\end{tabular}
\end{table}


\begin{table}[h!]
\caption{\\Whole image binary classification (S vs V) Recall percentages obtained from 5-fold cross-validation.}
\label{tab:recall}
\centering
\begin{tabular}{|c|c|c|c|c|c|}
\hline
\textbf{Classifier}                                                        & \textbf{R 1} & \textbf{R 2} & \textbf{R 3} & \textbf{R 4} & \textbf{R 5} \\ \hline
 A                                 & 88   & 84   & 72     & 84 & 80      \\ \hline
 B                                  & 60    & 60   & 56  & 80  & 64              \\ \hline
\begin{tabular}[c]{@{}c@{}}\textbf{Our ensemble}\\  \textbf{classifier}\end{tabular} & \textbf{92}        & \textbf{100}                    & \textbf{92}  & \textbf{100}& \textbf{92}           \\ \hline
\end{tabular}
\end{table}


\begin{table}[h!]
\caption{\\Whole image binary classification (S vs V) F1-score percentages obtained from 5-fold cross-validation.}
\label{tab:F1}
\centering
\begin{tabular}{|c|c|c|c|c|c|}
\hline
\textbf{Classifier}                                                        & \textbf{R 1} & \textbf{R 2} & \textbf{R 3} & \textbf{R 4} & \textbf{R 5} \\ \hline
 A                                                               & 89.81                   & 87.49                    & 83.72                 & 84 & 83.35       \\ \hline
 B                                                               & 62.49                  & 63.83     & 65.12 & 81.61 &69.56              \\ \hline
\begin{tabular}[c]{@{}c@{}}\textbf{Our ensemble}\\ \textbf{classifier}\end{tabular} & \textbf{93.86}                   & \textbf{94.34}                    & \textbf{95.83}  & \textbf{92.58}& \textbf{92}           \\ \hline
\end{tabular}
\end{table}


\begin{table}[h!]
\caption{\\Whole image 3-class classification (S vs V vs D) accuracy percentages obtained from 5-fold cross-validation.}
\label{tab:acc3class}
\centering
\begin{tabular}{|c|c|c|c|c|c|}
\hline
\textbf{Classifier}                                                        & \textbf{R 1} & \textbf{R 2} & \textbf{R 3} & \textbf{R 4} & \textbf{R 5} \\ \hline
 A                                                               & 79.1                   & 76.7                  & 82.6                 & 83.7 & 81.4       \\ \hline
 B                                                               & 68.6                  & 55.8     & 64 & 72.1 &61.6              \\ \hline
\begin{tabular}[c]{@{}c@{}}\textbf{Our ensemble}\\ \textbf{classifier}\end{tabular} & \textbf{84.9}                   & \textbf{81.4}                    & \textbf{88.4}  & \textbf{91.9}& \textbf{91.9}           \\ \hline
\end{tabular}
\end{table}


 \begin{table}[h!]
 \small
 \setlength{\arrayrulewidth}{0.25mm}
\caption{\\Whole image 3-class classification (S vs V vs D) precision percentages obtained from 5-fold cross-validation.}
\label{tab:precision3class}
\centering
\begin{tabular}{|c|c|c|c|c|c|c|}\hline
\textbf{Classifier} & \textbf{Class} & \textbf{R 1} & \textbf{R 2} & \textbf{R 3} & \textbf{R 4} & \textbf{R 5}  \\
\noalign{
\hrule height 1.5pt
} 
\multirow{3}{*}{ \textbf{A}} & D & 80 & 80 & 83.3 & 81.3 & 78.8 \\\cline{2-7}
    & S & 73.9 & 76.2 & 84.2 & 94.7 & 80 \\\cline{2-7}
    & V & 81.8 & 74.3 & 81.1 & 80 & 84.8 \\ \noalign{
\hrule height 1.5pt
} 
    \multirow{3}{*}{\textbf{B}} & D & 63.6 & 70.6 & 68 & 67.7 & 56.7 \\\cline{2-7}
    & S & 65 & 41.4 & 44 & 66.7 & 63.2 \\\cline{2-7}
    & V & 75.8 & 60 & 75 & 79.4 & 64.9 \\
    \noalign{
\hrule height 1.5pt
} 
  \multirow{3}{*}{  \begin{tabular}[c]{@{}c@{}} \textbf{ Our ensemble} \\ \textbf{classifier} \\ 
  \end{tabular}} & D & 81.8 & 92.6 & 84.4 & 89.7 & 85.3 \\\cline{2-7}
    & S & 85.7 & 70.4 & 90.5 & 100 & 91.7 \\\cline{2-7}
    & V & 87.5 & 81.3 & 90.9 & 87.9 & 100 \\
    \noalign{
\hrule height 1.5pt
} 
\end{tabular}
\end{table}
 

 
 \begin{table}[h!]
 \small
 \setlength{\arrayrulewidth}{0.25mm}
\caption{\\Whole image 3-class classification (S vs V vs D) recall percentages obtained from 5-fold cross-validation.}
\label{tab:recall3class}
\centering
\begin{tabular}{|c|c|c|c|c|c|c|}\hline
\textbf{Classifier} & \textbf{Class} & \textbf{R 1} & \textbf{R 2} & \textbf{R 3} & \textbf{R 4} & \textbf{R 5}  \\
\noalign{
\hrule height 1.5pt
} 
\multirow{3}{*}{ \textbf{A}} & D & 80 & 80 & 83.3 & 86.7 & 86.7 \\\cline{2-7}
    & S & 68 & 64 & 64 & 72 & 64 \\\cline{2-7}
    & V & 87.1 & 83.9 & 96.8 & 90.3 & 90.3 \\ \noalign{
\hrule height 1.5pt
} 
    \multirow{3}{*}{\textbf{B}} & D & 70 & 40 & 56.7 & 70 & 56.7 \\\cline{2-7}
    & S & 52 & 48 & 44 & 56 & 48 \\\cline{2-7}
    & V & 80.6 & 77.4 & 87.1 & 87.1 & 77.4 \\
    \noalign{
\hrule height 1.5pt
} 
  \multirow{3}{*}{  \begin{tabular}[c]{@{}c@{}} \textbf{Our ensemble} \\ \textbf{classifier} \\
  \end{tabular}} & D & 90 & 83.3 & 90 & 86.7 & 96.7 \\\cline{2-7}
    & S & 72 & 76 & 76 & 96 & 88 \\\cline{2-7}
    & V & 90.3 & 83.9 & 96.8 & 93.5 & 90.3 \\
    \noalign{
\hrule height 1.5pt
} 
\end{tabular}
\end{table}
 

 
 \begin{table}[h!]
 \small
 \setlength{\arrayrulewidth}{0.25mm}
\caption{\\Whole image 3-class classification (S vs V vs D) F1-score percentages obtained from 5-fold cross-validation.}
\label{tab:fscore3class}
\centering
\begin{tabular}{|c|c|c|c|c|c|c|}\hline
\textbf{Classifier} & \textbf{Class} & \textbf{R 1} & \textbf{R 2} & \textbf{R 3} & \textbf{R 4} & \textbf{R 5}  \\
\noalign{
\hrule height 1.5pt
} 
\multirow{3}{*}{ \textbf{A}} & D & 80 & 80 & 83.3 & 83.91 & 82.56 \\\cline{2-7}
    & S & 70.82 & 69.56 & 72.72 & 81.80 & 71.11 \\\cline{2-7}
    & V & 84.36 & 78.80 & 88.25 & 84.83 & 87.46 \\ \noalign{
\hrule height 1.5pt
} 
    \multirow{3}{*}{\textbf{B}} & D & 66.64 & 51.06 & 61.83 & 68.83 & 56.7 \\\cline{2-7}
    & S & 57.77 & 44.45 & 44 & 60.88 & 54.56 \\\cline{2-7}
    & V & 78.12 & 67.59 & 80.59 & 83.07 & 70.60 \\
    \noalign{
\hrule height 1.5pt
} 
  \multirow{3}{*}{  \begin{tabular}[c]{@{}c@{}} \textbf{Our} \\ \textbf{ensemble}\\\textbf{classifier} \\
  \end{tabular}} & D & 85.70 & 87.70 & 87.11 & 88.17 & 90.64 \\\cline{2-7}
    & S & 78.25 & 73.09 & 82.61 & 97.95 & 89.81 \\\cline{2-7}
    & V & 88.87 & 82.57 & 93.75 & 90.61 & 94.90 \\
    \noalign{
\hrule height 1.5pt
} 
\end{tabular}
\end{table}

\begin{figure*}[htp]
\centering
\includegraphics[width=2.7cm,height=2.7cm]{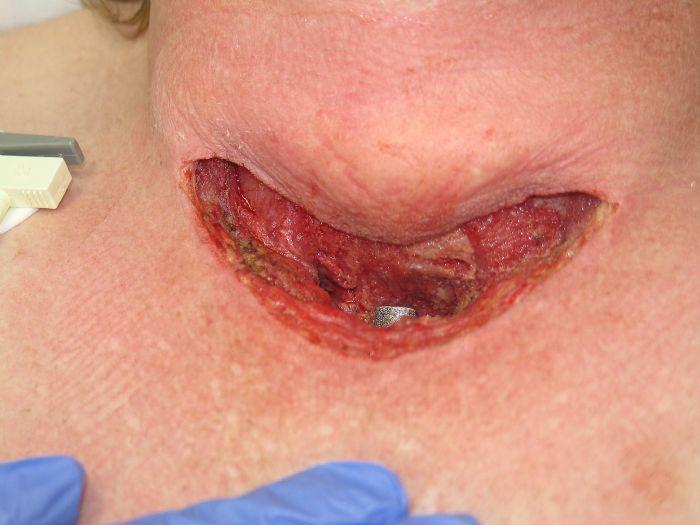}\quad
\includegraphics[width=2.7cm,height=2.7cm]{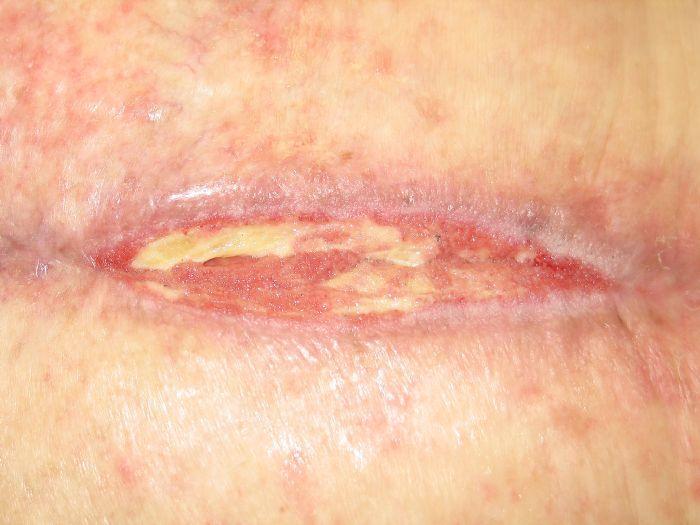}\quad
\includegraphics[width=2.7cm,height=2.7cm]{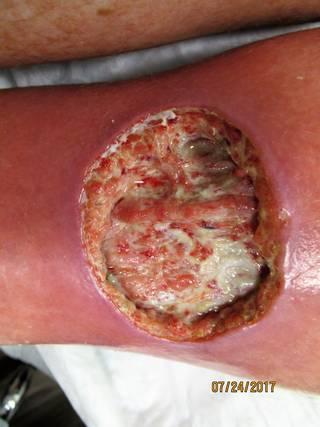}\quad
\includegraphics[width=2.7cm,height=2.7cm]{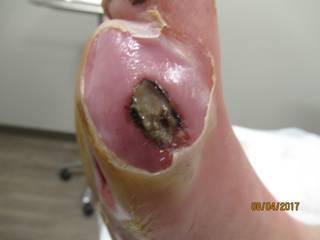}

\medskip

\includegraphics[width=2.7cm,height=2.7cm]{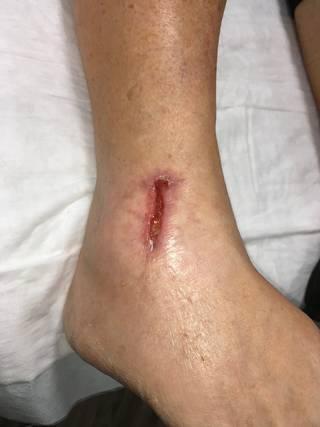}\quad
\includegraphics[width=2.7cm,height=2.7cm]{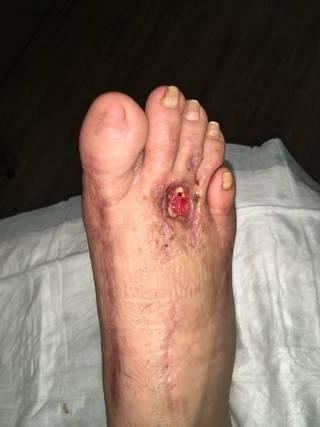}\quad
\includegraphics[width=2.7cm,height=2.7cm]{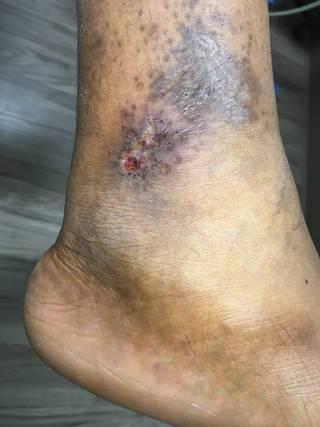}\quad
\includegraphics[width=2.7cm,height=2.7cm]{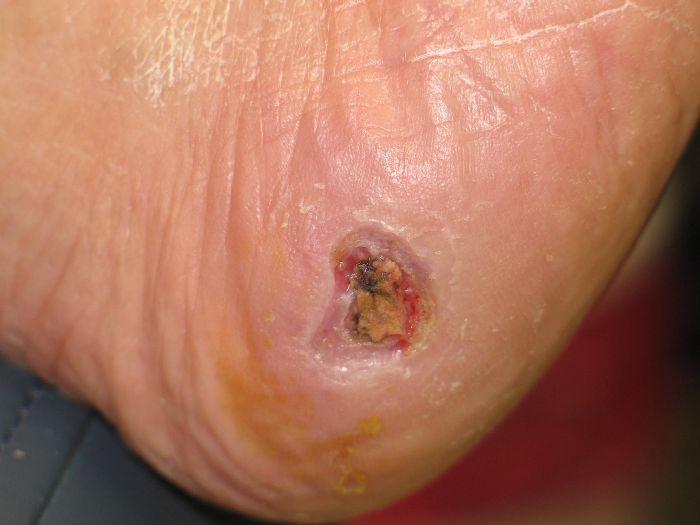}

\caption{Some of the miss-classified samples. The top row shows the samples miss-classified by Classifier B, and the bottom row displays the images which were wrongly classified by Classifier A.}
\label{fig:miss}
\end{figure*}

 Regarding the reported results, we find that our proposed ensemble classifier beats both Classifiers A and B with displaying better classification performance. This finding confirms our initial theory that combining two individual classifiers that one of them includes the patch-level information, and the other one contains the image-wise information will result in an stronger classifier with higher classification accuracy and better performance. Figure~\ref{fig:miss} presents some of the samples that were classified wrongly by Classifier A or B, but the proposed ensemble classifier assigned them to the correct class. The interesting observation was that the major part of the samples miss-classified by Classifier A, were the images in which the wound consists a small proportion of the entire image. On the other hand, Classifier B missed the samples in which the wound occupies a large part of the image. These observations show that how the two weak classifiers cooperate to fix each other's shortcomings which resulted in producing a superior classifier. Indeed, having objects from different scales in the dataset has always been one of the challenges in deep learning-based tasks as discussed in several studies~\cite{van2017learning, hesamian2019deep, zhao2019object}. Specifically, in the field of wound care it is not guaranteed to take high quality images from an optimal view point and desired distance to the wound surface because of medical concerns like infection control as well as the patient's easement~\cite{goyal2018dfunet}. Our results show that the Classifier B can overcome partially the scale problem for those images which were taken from a further distance to the wound. It of course loses this power for the photos zoomed on the wound area.  \par


\section{Conclusion}
In this paper, we proposed an end-to-end ensemble deep CNN-based classifier for classification of wound images into multiple classes based on the type of the wound. To the best of our knowledge, our proposed classifier is the first model that classifies the wound images into more than two types. We initially designed patch classifiers with fine-tuned AlexNet architecture to efficiently classify the wound patches into different wound types. Influence of different wound types on the classification accuracy was investigated by running numerous experiments and testing different combinations of the wound types. For image-wise classification task, for each input image, first a feature vector created from the designed patch classifier and another AlexNet that was trained on the whole images. Then the feature vector fed into an MLP to obtain an ensemble image-wise classifier with a higher accuracy and better performance. The results show that our proposed ensemble classifier can be used successfully as a decision support system in wound image classification tasks to assist the physicians in related clinical applications. We have made available the dataset we used for the current research. As a future study, we plan to improve the performance and classification accuracy of our proposed method by trying different combinations of the patch-wise and image-wise classifiers. Besides, testing the proposed approach to classify the images into more classes by working on a larger dataset of wound images would be one of the subsequent steps of this research.


\section*{Acknowledgment}
This work is partially supported by the Discovery and Innovation Grant (DIG)
award and the Catalyst Grant Program at The University of Wisconsin-Milwaukee.

\label{S:6}







\bibliographystyle{IEEEtran.bst}

\bibliography{sample.bib}







\end{document}